\documentclass[journal]{IEEEtran}
\usepackage{cite}
\usepackage{mathtools}

\ifCLASSINFOpdf
   \usepackage[pdftex]{graphicx}
\else
   \usepackage[dvips]{graphicx}
\fi
\usepackage{amsmath}
\usepackage{amssymb}
\usepackage[linesnumbered,ruled]{algorithm2e} 
\usepackage{algpseudocode}
\usepackage{txfonts}
\usepackage{newtxmath}
\usepackage{setspace}
\usepackage{embrac}

\usepackage{array}

\ifCLASSOPTIONcompsoc
 \usepackage[caption=false,font=normalsize,labelfont=sf,textfont=sf]{subfig}
\else
 \usepackage[caption=false,font=footnotesize]{subfig}
\fi
\usepackage{makecell}
\usepackage{hhline}
\usepackage{booktabs}
\usepackage{multirow}
\usepackage[normalem]{ulem}
\useunder{\uline}{\ul}{}

\usepackage{stfloats}
\begin{document}
%
\title{Learning Polar Encodings for Arbitrary-Oriented \\ Ship Detection in SAR Images}
%
%
%

\author{Yishan~He,
        Fei~Gao,
        Jun~Wang,
        Amir~Hussain,
        Erfu~Yang,
        Huiyu~Zhou
\thanks{This research was funded by the National Natural Science Foundation of China, Grant Nos. 61771027, 61071139, 61471019, 61501011 and 61171122. Professor A. Hussain was supported by the U.K. Engineering and Physical Sciences Research Council (EPSRC) Grant No. EP/M026981/1. H. Zhou was supported by the U.K.EPSRC under Grant EP/N011074/1, Royal Society-Newton Advanced Fellowship under Grant NA160342 and the European Union’s Horizon 2020 research and innovation program under the Marie Sklodowska Curie Grant Agreement No. 720325. \textit{(Corresponding author: Fei Gao.)}}
\thanks{Y. He, F. Gao, and J. Wang are with the School of Electronic and Information Engineering, Beihang University, Beijing 100191, China (e-mail: heyishan@buaa.edu.cn; feigao2000@163.com; wangj203@buaa.edu.cn).}
\thanks{A. Hussain is with the Cyber and Big Data Research Laboratory, Edinburgh Napier University, Edinburgh EH11 4BN, U.K., and also with the Taibah Valley, Taibah University, Medina 30001, Saudi Arabia (e-mail: A.Hussain@napier.ac.uk).}
\thanks{E. Yang is with Space Mechatronic Systems Technology Laboratory, Department of Design, Manufacture and Engineering Management, Strathclyde Space Institute, University of Strathclyde, Glasgow G1 1XJ, U.K. }
\thanks{H. Zhou is with the Department of Informatics, University of Leicester, Leicester LE1 7RH, U.K. (e-mail: hz143@leicester.ac.uk).}}

%
%

\markboth{IEEE JOURNAL OF SELECTED TOPICS IN APPLIED EARTH OBSERVATIONS AND REMOTE SENSING}%
{Shell \MakeLowercase{\textit{et al.}}: Learning Polar Encodings for Arbitrary-Oriented Ship Detection in SAR Images}
%



\maketitle

\begin{abstract}
Common horizontal bounding box (HBB)-based methods are not capable of accurately locating slender ship targets with arbitrary orientations in synthetic aperture radar (SAR) images. Therefore, in recent years, methods based on oriented bounding box (OBB) have gradually received attention from researchers. However, most of the recently proposed deep learning-based methods for OBB detection encounter the boundary discontinuity problem in angle or key point regression. In order to alleviate this problem, researchers propose to introduce some manually set parameters or extra network branches for distinguishing the boundary cases, which make training more difficult and lead to performance degradation. In this paper, in order to solve the boundary discontinuity problem in OBB regression, we propose to detect SAR ships by learning polar encodings. The encoding scheme uses a group of vectors pointing from the center of the ship target to the boundary points to represent an OBB. The boundary discontinuity problem is avoided by training and inference directly according to the polar encodings. In addition, we propose an Intersect over Union (IOU) -weighted regression loss, which further guides the training of polar encodings through the IOU metric and improves the detection performance. Experiments on the Rotating SAR Ship Detection Dataset (RSSDD) show that the proposed method can achieve better detection performance over other comparison algorithms and other OBB encoding schemes, demonstrating the effectiveness of our method.
\end{abstract}


%
\IEEEpeerreviewmaketitle

\section{Introduction}
%
%
%
%
\IEEEPARstart{S}{hip} detection in synthetic aperture radar (SAR) images is an important branch in SAR image interpretation. It can be widely applied in many fields, such as harbor monitoring, fishery monitoring, maritime traffic monitoring, intelligence acquisition and so on \cite{mazzarella2015sar,ao2018detection,wang2016ship,zhao2020attention,sun2021pbnet}. Therefore, it has attracted much attention in recent years. Traditional SAR ship detection algorithms usually include multi-steps as follows: (1) sea-land segmentation; (2) image preprocessing; (3) candidate region extraction and (4) false alarm rejection. Based on the pipeline, researchers have proposed a variety of methods, which can be mainly classified as threshold-based \cite{dai2016modified,pappas2018superpixel}, saliency-based \cite{gao2017visual,wang2012hierarchical}, hand-crafted feature-based \cite{leng2018area,lin2018ship} and statistical modeling-based methods \cite{gao2016scheme, pan2018ship}. With the increase of SAR data amount and resolution, these methods face difficulties in meeting the practical demands in terms of accuracy, robustness and speed\cite{yue2019novel,gaonew,zhang2019novel,wang2019fmssd,sun2020sraf}. It is mainly due to their complex detection flows and high dependence on prior knowledge such as specific statistical distribution modeling and manually-designed features \cite{zhang2019depthwise,cui2019dense,deng2019learning,fan2019ship,cui2020ship}. Hence, it is urgent to develop smarter and more automated SAR ship detection methods.

Recently, with the development of deep learning theories and the substantial improvement of the hardware, deep convolutional neural network (DCNN)-based algorithms have achieved great success in computer vision fields, such as target detection, recognition, segmentation, tracking and so on \cite{zhao2019object,deng2009imagenet,long2015fully,krizhevsky2012imagenet,szegedy2013deep,dai2016r,sun2020bas, sun2020deep}. In the field of target detection, Ren et al. \cite{ren2015faster} extracted candidate target regions through region proposal network (RPN) in Faster-RCNN and conducted end-to-end training to achieve high detection accuracy and speed. To achieve higher efficiency, researchers put forward to use DCNN to directly regress the target locations without extracting candidate regions, such as Single-Shot multi-box Detector (SSD) \cite{liu2016ssd}, RetinaNet \cite{lin2017focal}, YOLO \cite{redmon2016you, redmon2018yolov3}, etc. In addition, methods based on key point detection have received much concern lately \cite{law2018cornernet,zhou2019objects,yang2019reppoints,tian2019fcos}. For instance, Law et al. \cite{law2018cornernet} proposed to locate targets by regressing the upper left and lower right corners of the bounding boxes; CenterNet by Xing et al. \cite{zhou2019objects} detected targets by locating the center points and the length and width of the bounding boxes. In the task of ship detection in SAR images, the DCNN-based algorithms also achieve great performance. For example, Deng et al. \cite{deng2019learning} trained DCNN-based ship detector from scratch on SAR dataset by introducing dense blocks and new training losses; Liu et al. \cite{liu2017sar} combined pyramid features extracted by DCNN into sea-land segmentation and the ship detection process; Gao et al. \cite{gao2020anchor} improved CenterNet by attention mechanism and feature reuse strategy to achieved good performance in SAR ship detection. 

\begin{figure}[t]
\centering  
\subfloat[]{  
\includegraphics[scale=0.7]{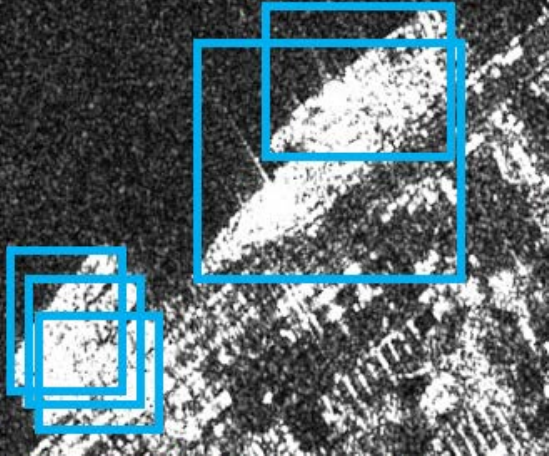}}
\hspace{2pt} 
\subfloat[]{  
\includegraphics[scale=0.7]{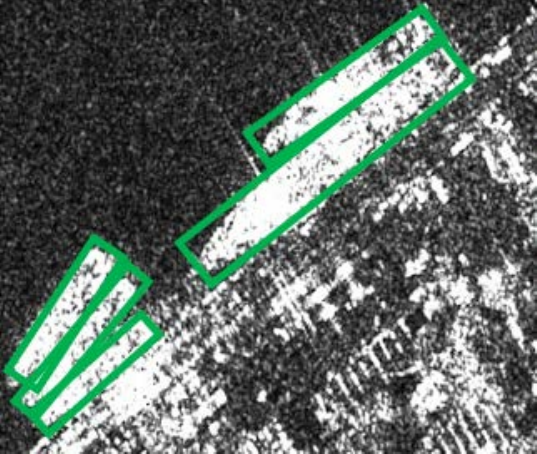}}  
\caption{Comparison between the HBB-based detection and OBB-based detection. (a) Results of HBB-based detection; (b) Results of OBB-based detection.}  
\label{fig1}
\end{figure}

The above methods all adopt the horizontal bounding box (HBB) for target localization. However, HBB-based methods suffer certain difficulties. In SAR images, ship targets are slender, arbitrary-oriented and sometimes densely distributed. In these cases, HBB-based methods may fail. As shown in Fig. \ref{fig1}(a), the HBBs of the ship targets are overlapped and hard to be distinguished from each other. To overcome such difficulty, researchers turned to oriented bounding box(OBB)-based methods. As shown in Fig. \ref{fig1}(b), the OBB-based detection results can clearly indicate the shape and orientation of the ship targets, which effectively avoid overlapping and give more accurate results.

OBB-based detection methods can be mainly categorized as three types: (1) anchor-based methods; (2) angle prediction-based methods; and (3) key point regression-based methods. Anchor-based methods locate targets by predicting the errors between the preset anchors and the actual bounding boxes of the targets. For example, Xia et al. \cite{xia2018dota} proposed FRCNN-OBB algorithm, which utilized Faster-RCNN-based framework to regress the errors between the OBBs and horizontal anchors. Ding et al. \cite{ding2019learning} trained DCNN to transform horizontal region of interest(H-ROI) into rotated region of interest(R-ROI), so as to improve the feature extraction ability of the network for arbitrary-oriented targets. For angle prediction-based methods, the OBB of the target is identified by these parameters: the location of the target, the length, width and rotation angle of the bounding box. For instance, Hu et al. \cite{zhu2020adaptive} put forward to use two-dimensional periodic vectors to represent the angle of OBB, and a length-independent intersection over union metric is proposed to guide network training. Yang et al. \cite{yang2020arbitrary} predicted the rotation angle of OBB by classification rather than regression and designed a periodic loss function, so as to alleviate the boundary discontinuity for OBB regression. Key point regression-based methods represent OBB as key points. For example, Yi et al. \cite{yi2021oriented} denoted target OBBs with four vectors distributed in different quadrants of Cartesian coordinates. Xu et al. \cite{xu2020gliding} first predicted the horizontal enclosing rectangle of the target OBB, then regressed the distances between the vertices of the enclosing rectangle and the vertices of the OBB to position the OBB. Zhao et al. \cite{zhao2020polardet} regressed the polar coordinates of the four vertices of the target OBB to achieve OBB localization. Fu et al. \cite{fu2020point} detected the OBB of the target by utilizing fully convolutional networks to locate a group of points distributed evenly inside the OBB.

In the field of ship detection in SAR images, OBB-based methods have also been widely studied. For example, Wang et al. \cite{wang2018simultaneous} proposed an improved SSD network, where the angle information was utilized to further encode the orientation of the ship targets. Chen et al. \cite{chen2020r2fa} built a feature alignment module to extract the features of the ship targets more accurately 
based on oriented anchors. Chen et al. \cite{chen2019msarn} designed multi-layer anchors and rotation non-maximum suppression post-processing to improve the detection performance for oriented ship targets. Pan et al. \cite{pan2020msr2n} used rotating region proposal network (RRPN) to extract candidate target regions, and then multi-layer cascade network was employed to fine tune the OBB detection results. An et al. \cite{an2019drbox} proposed multi-layer anchor settings and a new encoding scheme to compute the errors between anchors and the predicted OBBs, so as to alleviate the boundary discontinuity problem in OBB prediction. 

The fore-mentioned OBB-based methods encounter the boundary discontinuity problem in varying degrees. The causes of the problem can be attributed to two sides, the periodicity of angle (POA) and the exchangeability of edge (EOE) \cite{zhu2020adaptive,xu2020gliding,yang2020dense}. The boundary discontinuity problem leads to mismatching between annotations and predictions during the training stage, causing performance degradation. To deal with the problem, researchers propose different approaches. For example, Yi et al. \cite{yi2021oriented} and Xu et al. \cite{xu2020gliding} first set a fixed threshold manually for distinguishing HBB from OBB, and added a network branch to classify the target bounding box as HBB or OBB. Different regression rules were adopted for different kinds of target bounding boxes. However, this kind of solution leads to more complex network structures, difficulty on tuning the parameters and decrease of the network convergence. Hu et al. \cite{zhu2020adaptive} used periodic loss function for angle regression of OBB to reduce the negative influence of the POA on network training. However, this approach still needs to manually set parameters to define when the boundary discontinuity problem occurs. Yang et al. \cite{yang2020arbitrary} transformed the task of angle regression into angle classification, so as to avoid the boundary discontinuity problem. However, angle quantization brings unacceptable computation burden on the classification task, leading to poor real-time performance. 

To provide a direct and effective solution for the boundary discontinuity problem, in this paper, we propose to detect ship targets in SAR images based on polar encodings. Through the polar encoding and decoding process, the boundary discontinuity problem can be naturally addressed. To be specific, we encode the OBB of the ship target by sampling a group of ordered boundary points on the OBB. In this way, the ground truth is in one to one correspondence with the prediction at each fixed angle, which prevents ambiguity in training. Furthermore, the function of the sampling distance to the sampling angle is periodic, which guarantees the continuity in boundary cases. In addition, to further guide the training of polar vectors and improve the detection performance, we propose to use Intersect over Union(IOU) metric to weight the regression loss. The experimental results on the Rotating SAR Ship Detection Dataset (RSSDD) are given to verify the effectiveness of the proposed polar encoding scheme and the IOU-weighted regression loss function. The comparison results with other OBB-based detection methods demonstrate that our method outperforms other comparison methods, achieving better detection results.

The rest of the paper is organized as follows: Section II describes the boundary discontinuity problem and our proposed method as solution in detail. The experimental results on RSSDD dataset are given in Section III. Section IV presents discussions. Section V concludes the paper.


\section{Proposed Method}

\begin{figure*}[t]
\centering
\includegraphics[scale=0.7]{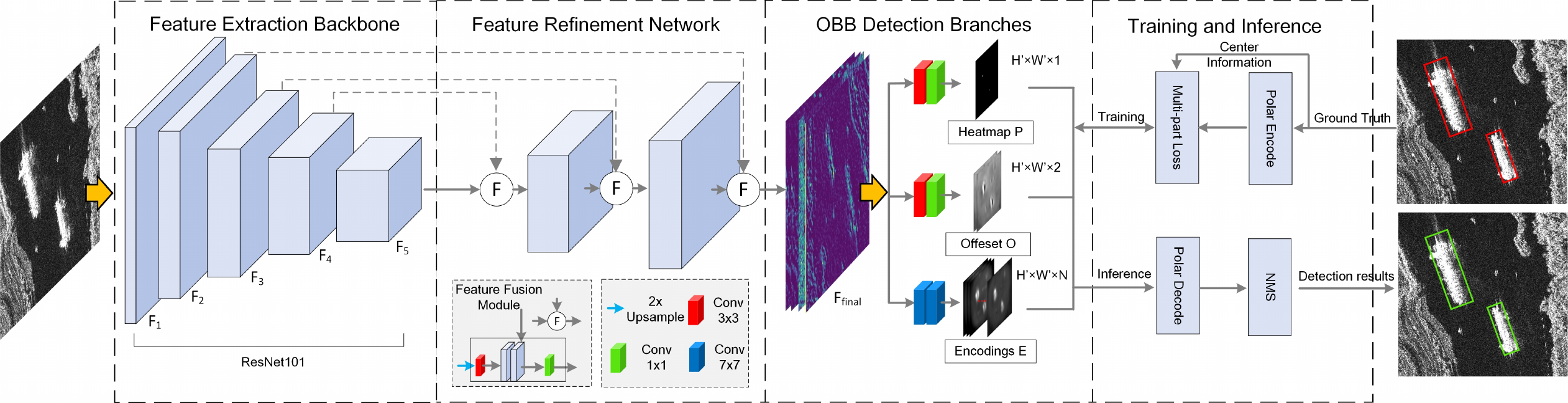}
\caption{The overall architecture of our method. The network structure can be divided into three parts: the feature extraction backbone, the feature refinement network and the OBB detection branches. They output the center heatmap P, the center offset map O and the encoding map E, respectively. In the training stage, the output of the detection branches is combined with the center information and the encoded OBB parameters for calculating the multi-part loss function. In the inference stage, the polar decoding process is proposed to obtain the detection results, and NMS algorithm is used for removing the duplicate targets.}
\label{fig2}
\end{figure*}

The overall architecture of our method is illustrated in Fig. \ref{fig2}. Firstly, the input SAR image is fed as input of the ResNet-101 feature extraction backbone \cite{he2016deep}, through which features of five different scales $ \{ F_{1}, F_{2}, F_{3}, F_{4}, F_{5} \} $ are obtained. As the shallowest feature $F_{1}$ contains little semantic information, $\{ F_{2}, F_{3}, F_{4}, F_{5} \} $ are chosen to be combined through feature fusion module, and the resolution of the final output feature $F_{final}$ is 1/4 as the input image. $F_{final}$ is then processed by three network branches, from which we can obtain the center heatmap $P$, the center offset map $O$ and the encoding map $E$, respectively. In the training stage, the multi-part loss function is calculated according to the center information of the ship targets and the polar encodings. The losses are combined to train the branches jointly. In the inference stage, the output of the branches is decoded through the polar decoding process. And the non-maximum suppression algorithm(NMS) is adopted to remove the duplicate detections and obtain the final detection results. In the polar encoding process, for each ship target, we sequentially sample the distances between the center point of the ship target and the boundary of the OBB every $\pi/N$ in the range of $[0,\pi)$. The sampled N values are combined to form an encoding vector. Due to the central symmetry of the OBB, the encoding vector can represent the shape of the whole OBB. In the polar decoding process, the center points of the ship targets are first extracted from $P$. Then the downsampling quantization errors are compensated in terms of the predictions from $O$. The OBBs of the ship targets are finally restored through the processes of extracting the polar vectors from $E$, converting the polar vectors into the boundary point sets, and finding the minimum bounding boxes of the point sets. 

In this section, we will first introduce the boundary discontinuity problem from OBB detection; then we will describe the network architecture and the specific process of polar encoding and decoding in detail; finally, we will introduce the loss functions for network training.

\subsection{The Boundary Discontinuity Problem}
\label{bdp}
For OBB-based methods, problems occur in the boundary cases where the predictions of the OBB parameters will change discontinuously. In particular, the boundary discontinuity problem can be attributed to two reasons: the periodicity of angle (POA) and the exchangeability of edge (EOE). Due to the POA, the angle parameter suffers discontinuity. For instance, the lower and upper bounds of the angle parameter denote basically the same orientation but their values differ greatly. The EOE refers to the problem that the order of the lengths or key points of the OBB will suddenly change in the boundary cases, leading to discontinuity. The discontinuity caused by POA and EOE will lead to a high loss value even if the OBBs from the prediction and the ground truth share high overlap, which is prone to cause the convergence problem.

\begin{figure}[t]
\centering  
\subfloat[]{  
\hspace{6pt} 
\includegraphics[scale=0.6]{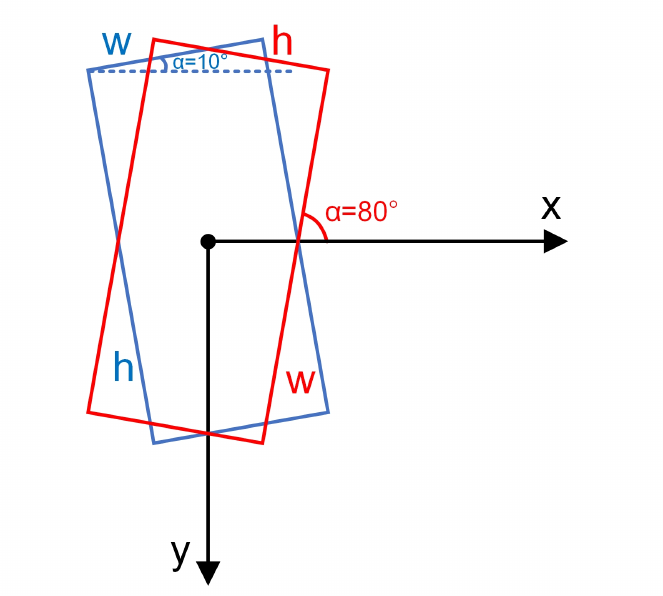}}
\hspace{2pt} 
\subfloat[]{  
\includegraphics[scale=0.6]{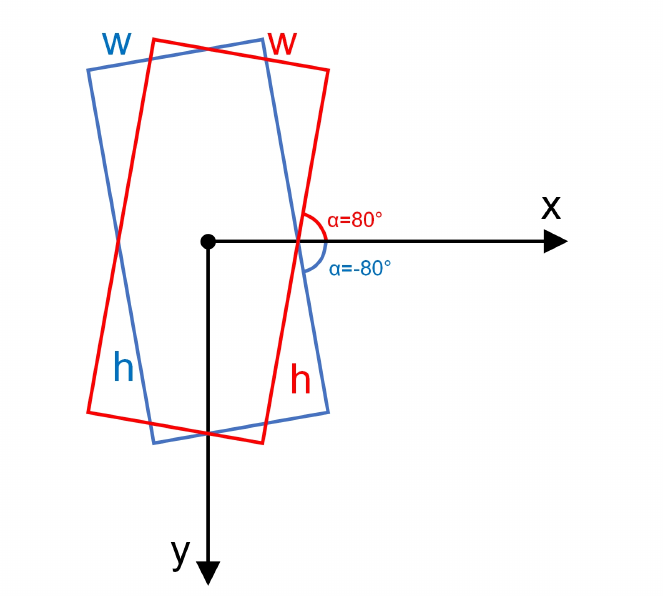}}
\vspace{2pt} 
\subfloat[]{  
\includegraphics[scale=0.6]{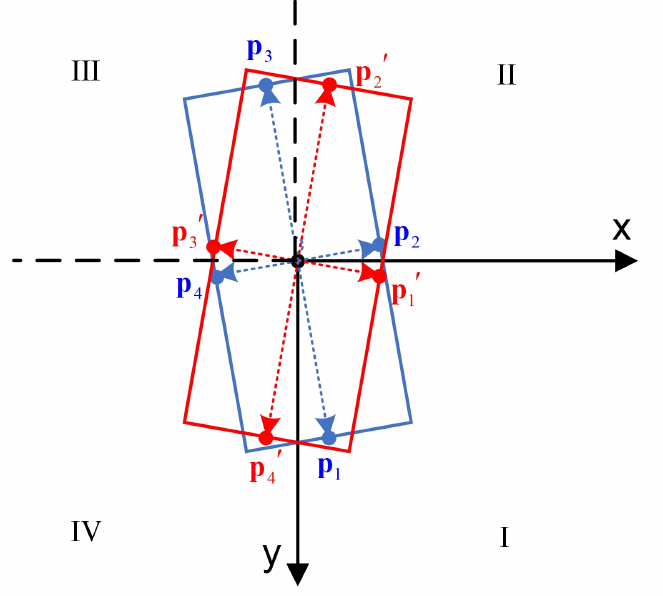}}
\hspace{2pt} 
\subfloat[]{  
\includegraphics[scale=0.6]{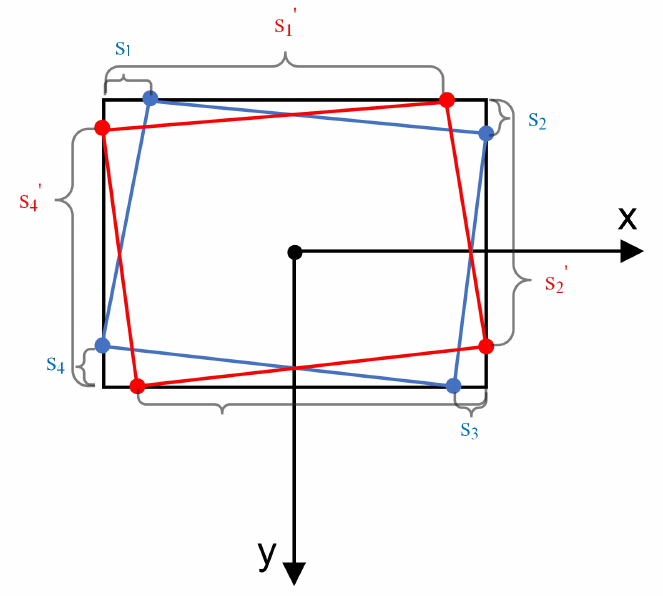}}  
\caption{The boundary cases for different OBB-based methods, the blue rectangle denotes the ground truth, the red rectangle represents the prediction. (a) the boundary case for 90°-based representation; (b) the boundary case for 180°-based representation; (c) the boundary case for \cite{yi2021oriented};(d) the boundary case for \cite{xu2020gliding}.}  
\label{fig3}
\end{figure}

For angle prediction-based methods, there are mainly two kinds of representations for OBB: 90°-based representation and 180°-based representation. In both cases, the OBB is determined by the center point, length, width and rotation angle $(ct,w,h,\alpha )$. For 90°-based representation, the rotation angle is defined as the angle from the x-axis counterclockwise to the first coincident edge of the OBB, the range of which is $[0, 90^{\circ})$. The length of the first edge that coincides is denoted as $w$. A typical boundary case is shown in Fig. \ref{fig3}(a). The predicted edges and angles are mismatch with the ground truth due to the POA and EOE. For the 180°-based representation, the rotation angle $\alpha$ is determined according to the angle from the x-axis to the long side of the OBB. The range of $\alpha$ is $(-90^{\circ},90^{\circ}]$. As shown in Fig. \ref{fig3}(b), in the boundary case, the edges of the predicted OBB and the ground truth correctly correspond to each other, but the angle suffers discontinuity because of the POA. In addition, the performance of the angle prediction-based methods is sensitive to angle prediction errors \cite{yang2020dense,zhao2020polardet}. As shown in Fig. \ref{fig4}, with large aspect ratios of the OBB, the small angle prediction errors will cause a rapid drop in IOU. It makes it difficult for the loss function to guide the training of the network

For key point regression-based methods, similar problems exist. Yi et al. \cite{yi2021oriented} expressed the target OBB as a group of midpoints from four edges of the OBB. The four midpoints are distributed in the four quadrants of the Cartesian Coordinates, respectively. In this way, the one to one correspondence between the prediction and the ground truth is established. The boundary case of this method is shown in Fig. \ref{fig3}(c). When the prediction and the ground truth share high overlap, the distance errors between the actual point set $({{\mathbf{p}}_{1}},{{\mathbf{p}}_{2}},{{\mathbf{p}}_{3}},{{\mathbf{p}}_{4}})$ and the predicted point set  $({\mathbf{p}'_{1}},{\mathbf{p}'_{2}},{\mathbf{p}'_{3}},{\mathbf{p}'_{4}})$ is large due to the EOE problem. Xu et al. \cite{xu2020gliding} determined the shape and orientation of OBB by regressing four distances between the four vertices of OBB and that of the HBB. As shown in Fig. \ref{fig3}(d), in the boundary case, the errors between the predicted distances $({{s}_{1}},{{s}_{2}},{{s}_{3}},{{s}_{4}})$ and the ground truth $({s'_{1}},{s'_{2}},{s'_{3}},{s'_{4}})$ are too large to indicate the actual overlap degree. In this paper, in order to avoid the fore-mentioned boundary discontinuity problems, we propose an encoding and decoding scheme for OBB-based ship detection in SAR images.

\subsection{The DCNN architecture of our method}
The overall DCNN structure of our method is illustrated in Fig. \ref{fig2}. It can be divided into three parts: the feature extraction backbone, the feature refinement network, and three branch networks for OBB detection. The detailed structure is described as follows:

\begin{figure}[t]
\centering  
\subfloat[]{  
\includegraphics[scale=0.3]{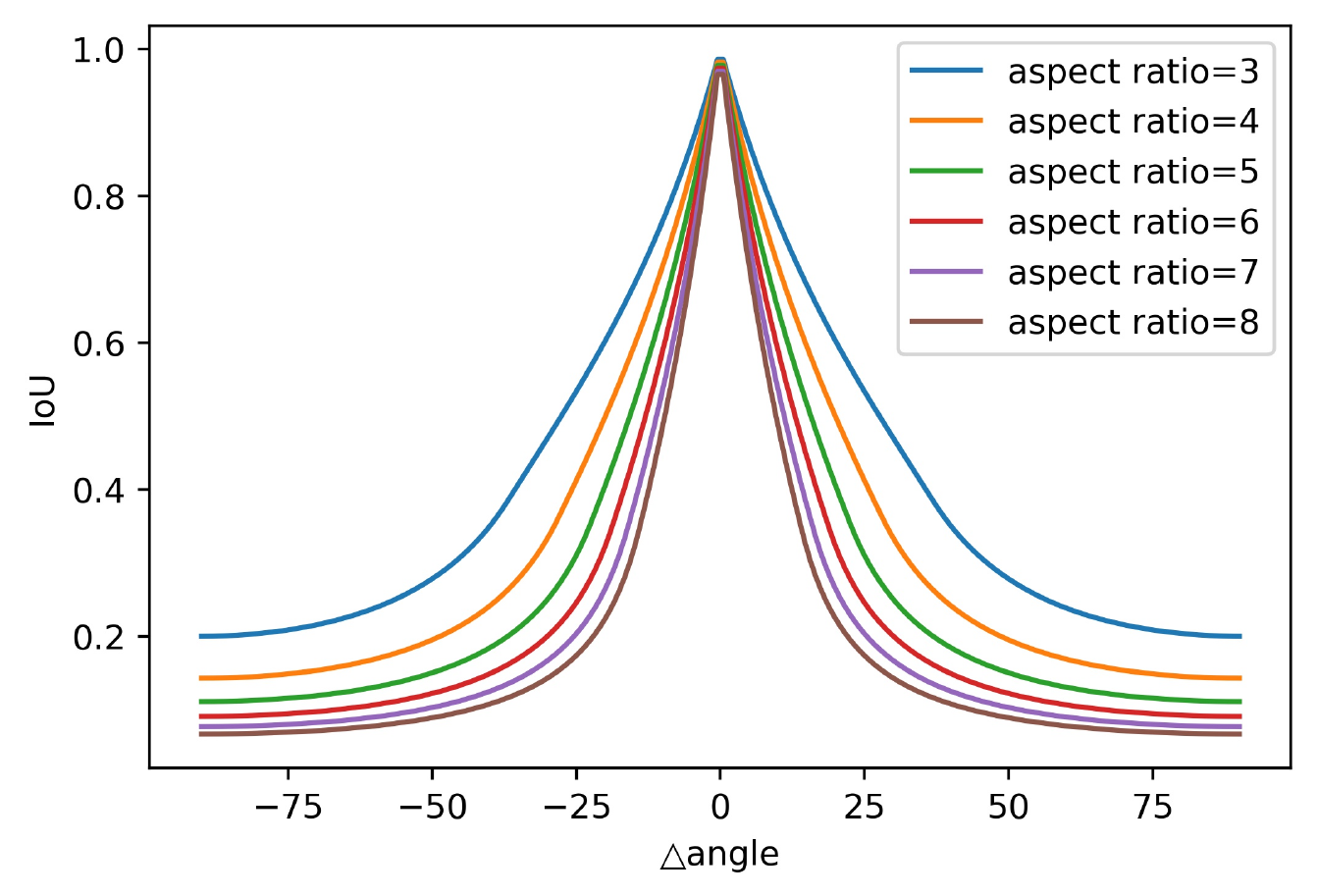}}
\hspace{2pt} 
\subfloat[]{  
\includegraphics[scale=0.3]{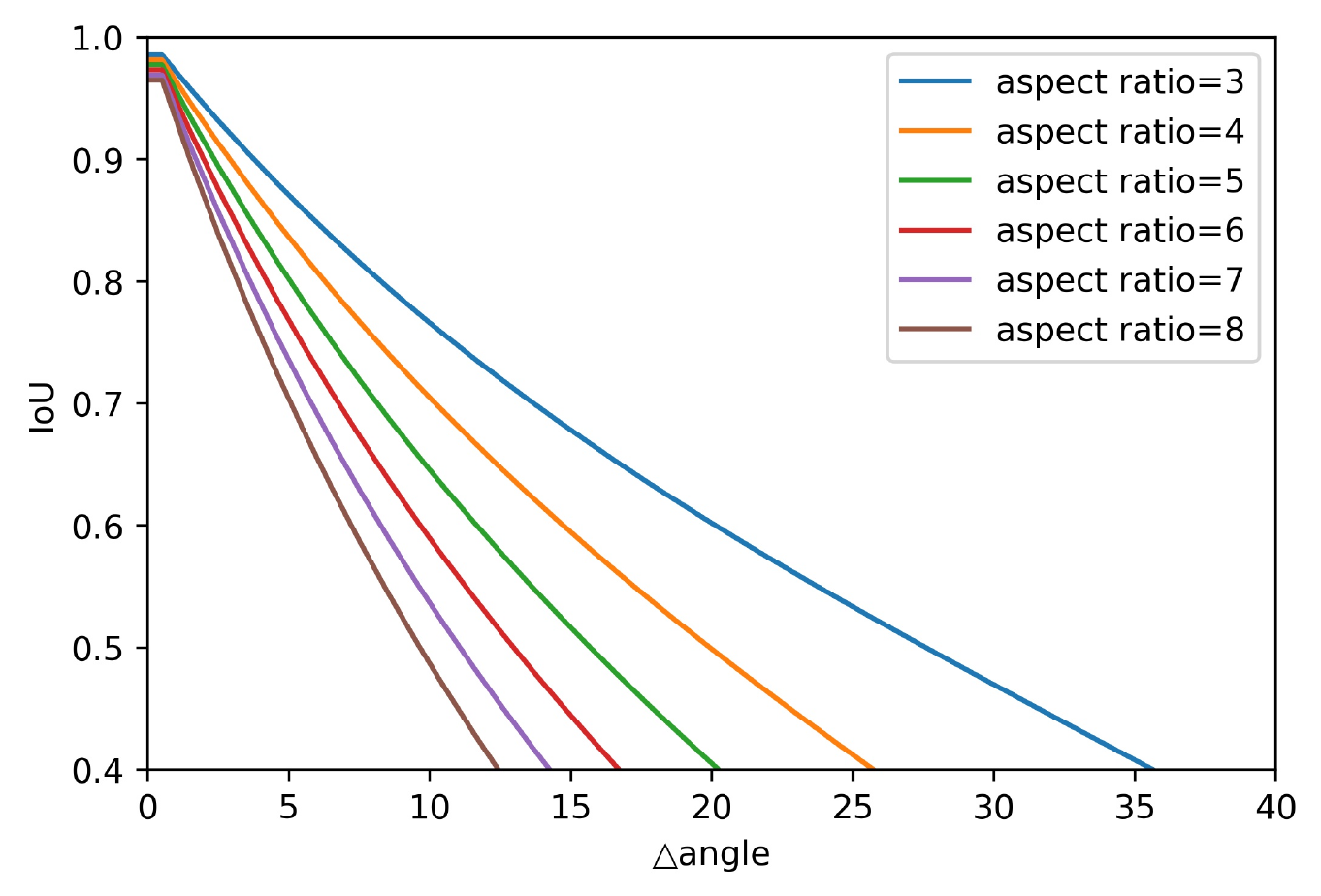}}  
\caption{The IOU sensitivity of the angle regression. (a) The relationship between the angle error and the detection IOU for OBBs with different aspect ratios; (b) A local view of (a).}  
\label{fig4}
\end{figure}

\subsubsection{Feature Extraction Backbone}
We adopt ResNet-101 \cite{he2016deep} as the feature extraction backbone. It consists of five convolutional stages. With the stage going deeper, the resolution of the features gradually decreases, the receptive field and the semantic information increase. Given the input SAR image $I\in {{\mathbb{R}}^{H\times W\times 3}}$, the feature extraction backbone generates five scales of features $ \{ F_{1}, F_{2}, F_{3}, F_{4}, F_{5} \} $. 
\subsubsection{Feature Refinement Network}
The deep features contain richer semantic information and larger receptive fields, which are suitable for detecting large ship targets. And the shallow features are of high resolution, which are helpful for detecting small targets. Therefore, different scales of features, $\{ F_{2}, F_{3}, F_{4}, F_{5} \} $ are fused in the upsamling process by the feature fusion module. As shown in Fig. \ref{fig2}, for two input features of different scales, the feature fusion module first upsamples the lower resolution feature and performs a 3×3 convolution. Then the upsampled feature are concatenate channel-wise with the high-resolution feature. The output feature is obtained by employing a 1×1 convolution for channel dimension reduction. The process of the feature fusion module can be represented as follows:
\begin{IEEEeqnarray}{l}\label{equ1}
    \begin{aligned}
        {{F}_{out}}&={{F}_{l}}\otimes {{F}_{h}} \\ &=Con{{v}_{1\times 1}}(Con{{v}_{3\times 3}}(Upsampl{{e}_{2\times }}({{F}_{l}}))\odot {{F}_{h}})
    \end{aligned}
\end{IEEEeqnarray}
where $F_{l}$ denotes the low-resolution input feature, $F_{h}$ is the high-resolution input feature, $\otimes$  denotes the feature fusion operation by the feature fusion module, $Upsampl{{e}_{2\times }}$  stands for the upsampling operation and $\odot$ represents the channel concatenation operation. $\{ F_{2}, F_{3}, F_{4}, F_{5} \} $ are fused through feature fusion module successively as follows:
\begin{IEEEeqnarray}{l}\label{equ2}
    {{F}_{final}}=(({{F}_{5}}\otimes {{F}_{4}})\otimes {{F}_{3}})\otimes {{F}_{2}}
\end{IEEEeqnarray}
where $F_{final}$ is the high-resolution feature output by the feature refinement network, whose size is $1 / 4$ of the input SAR image.

\subsubsection{OBB detection branches}
The feature output by the feature refinement network is then fed as input of three network branches, namely the center prediction branch, the offset regression branch and the encoding regression branch. The center prediction branch and the offset regression branch are both composed of a 3×3 convolution and a 1×1 convolution, which output the center heatmap $P\in {{\mathbb{R}}^{\tilde{H}\times \tilde{W}\times 1}}$ and the offset map $O\in {{\mathbb{R}}^{\tilde{H}\times \tilde{W}\times 2}}$, respectively. The process of the center prediction branch and the offset regression branch can be expressed as:
\begin{IEEEeqnarray}{l}\label{equ3}
    \begin{aligned}
      & P=C\text{on}{{\text{v}}_{1\times 1}}(Con{{v}_{3\times 3}}({{F}_{final}})) \\ 
     & O=C\text{on}{{\text{v}}_{1\times 1}}(Con{{v}_{3\times 3}}({{F}_{final}})) \\ 
    \end{aligned}
\end{IEEEeqnarray}

The encoding regression branch consists of two cascaded 7×7 convolutions and outputs the encoding map $E\in {{\mathbb{R}}^{\tilde{H}\times \tilde{W}\times N}}$, where $N$ is the number of encoding points. The process of the encoding regression branch can be represented by the following equation:
\begin{IEEEeqnarray}{l}\label{equ4}
    E=C\text{on}{{\text{v}}_{7\times 7}}(Con{{v}_{7\times 7}}({{F}_{final}}))
\end{IEEEeqnarray}

The detection results can be obtained by these three outputs through polar decoding process, which will be described in detail below.

\subsection{Polar Encoding}
\begin{figure*}[t]
\centering
\includegraphics[scale=0.95]{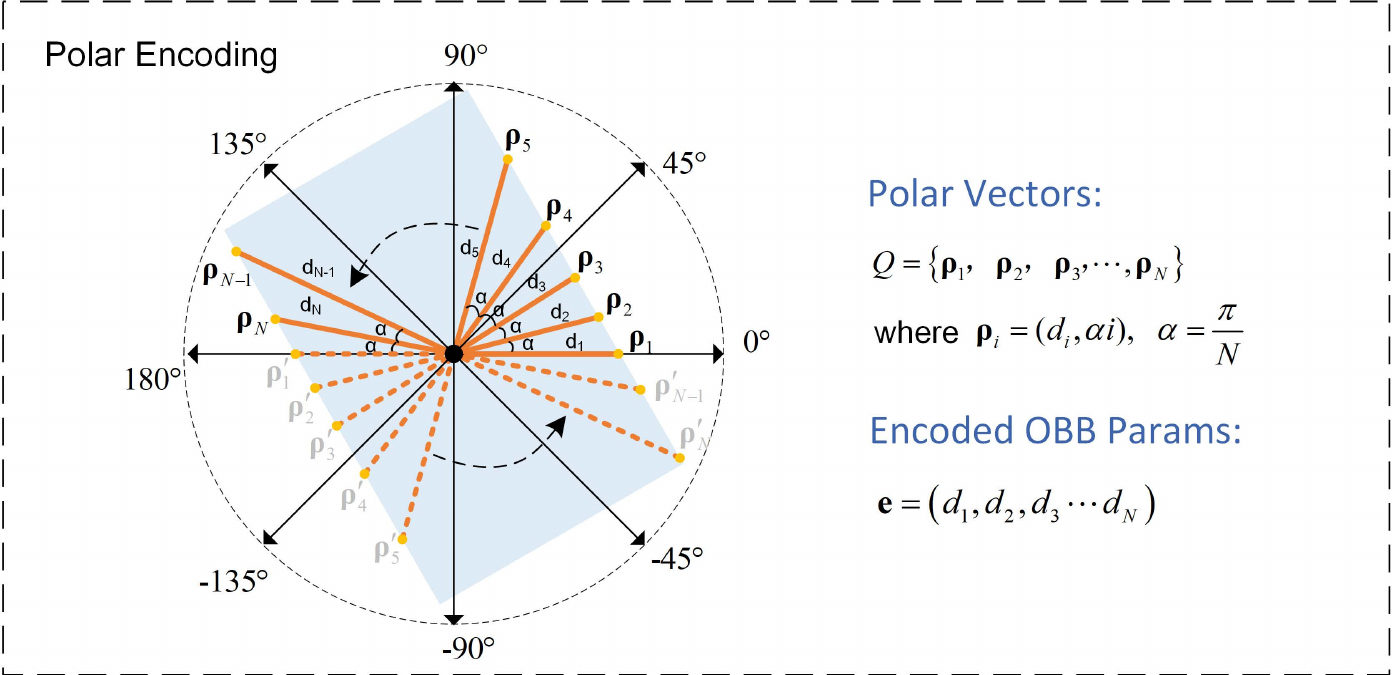}
\caption{The diagram of the polar encoding process. The polar vectors $Q$ are vectors pointing from the center to the boundary of the OBB. The lengths of the vectors are encoded as the OBB parameters $\boldsymbol{e}$.}
\label{fig6}
\end{figure*}

To avoid the boundary discontinuity problem, we propose to encode the OBB of ship target into a group of sequential values by using polar coordinates. The encoding diagram is shown in Fig. \ref{fig6}. The boundary points of the OBB are sampled at fixed angles. The distances between the boundary points and the ship center are collected as the OBB parameters, which can be predicted using DCNN.
The steps of the polar encoding are listed in Algorithm 1. For common methods, the OBB of the ship target is annotated by its four corners $\text{P}=\{{{\mathbf{p}}_{i}}|{{\mathbf{p}}_{i}}=({{x}_{i}},{{y}_{i}}),i=1,2,3,4\}$. The center point $\mathbf{c}$ can be calculated by $\mathbf{c}=({{x}_{c}},{{y}_{c}})=\sum\nolimits_{\text{i}=1}^{4}{{{\mathbf{p}}_{i}}}/4$. Then the corner vectors pointing from the center to the corners can be obtained by $V=\{{{\mathbf{v}}_{i}}|{{\mathbf{v}}_{i}}=({{x}_{i}}-{{x}_{c}},{{y}_{i}}-{{y}_{c}}),i=1,2,3,4\}$. For each vector ${{\mathbf{v}}_{i}}$ in $V$, calculate the vector angle ${{\alpha }_{i}}$:
\begin{IEEEeqnarray}{l}\label{equ5}
    {{\alpha }_{i}}={{(-1)}^{\text{u}({{y}_{i}}\text{-}{{y}_{c}})}}\cdot \arccos (\frac{{{x}_{i}}-{{x}_{c}}}{{{z}_{i}}})
\end{IEEEeqnarray}
where $\text{u}(\cdot )$ denotes the Heaviside function, ${{z}_{i}}=\sqrt{{{({{x}_{i}}-{{x}_{c}})}^{2}}+{{({{y}_{i}}-{{y}_{c}})}^{2}}}$ , the range of ${{\alpha }_{i}}$ is $(-\pi ,\pi ]$.

Next, sort the angles. The sorted angles divide $(-\pi ,\pi ]$ into four intervals: $({\alpha'_{1}},{\alpha'_{2}}],({\alpha'_{2}},{\alpha'_{3}}],({\alpha '_{3}},{\alpha '_{4}}],(-\pi ,{\alpha '_{1}}]\cup ({\alpha '_{4}},\pi ]$. In this way, the four edges of the OBB are in one to one correspondence with the four angle intervals, which is helpful for calculating the OBB parameters. According to the pre-set parameter $N$, boundary points are sampled from the OBB every $\pi/N$. The OBB parameters can be obtained by calculating the distances between the boundary points and the center point. Given the sampling angle $\theta_{i}$, assume that the polar coordinates of the two endpoints of the corresponding edge ${E'_{jk}}$ are $\left( \text r'_{j},\alpha'_{j} \right)$ and $\left( \text r'_{k},\alpha'_{k} \right)$, then the distance $d_{i}$ from the center point to sampled boundary point can be calculated by:
\begin{IEEEeqnarray}{l}\label{equ6}
    {{\text{d}}_{i}}\text{=}\frac{{{{{r}'}}_{j}}{{{{r}'}}_{k}}\sin ({{{{\alpha }'}}_{j}}-{{{{\alpha }'}}_{k}})}{{{{{r}'}}_{k}}\sin ({{\theta }_{i}}-{{{{\alpha }'}}_{k}})+{{{{r}'}}_{j}}\sin ({{{{\alpha }'}}_{j}}-{{\theta }_{i}})}
\end{IEEEeqnarray}

\begin{algorithm}[!t]\label{alg_1}
\setstretch{1}          
\caption{Polar Encoding}
\KwIn{$\{\left( x_{i},y_{i}\right) ,i=1,2,3,4\}$: Corner points of an OBB; $N$: the number of the encoding points. }

\KwOut{The OBB parameters $\boldsymbol{e}$.}
Calculate the collections of vectors $V$ pointing from the center to the corner points;\\
Calculate the angles of the vectors$ (\alpha_{1},\alpha_{2},\alpha_{3},\alpha_{4})$;\\
Sort the angles, the sorted angles $ (\alpha_{1}^{'},\alpha_{2}^{'},\alpha_{3}^{'},\alpha_{4}^{'})$ divides$(-\pi,\pi]$into four intervals, corresponding to the four edges of the OBB: $ (E_{12}^{'},E_{23}^{'},E_{34}^{'},E_{41}^{'})$; \\
\For{$i\leftarrow 1$ to $N$} {
    Calculate $\theta_{i}\leftarrow \pi i/N$;\\
    Find the corresponding edge ${E'_{jk}}:\left( {r'_{j}},{\alpha'_{j}} \right)\Leftrightarrow \left( {r'_{k}},{\alpha'_{k}} \right)$ of the OBB for $\theta_{i}$;\\
    Calculate the distance corresponding to $\theta_{i}$:
        $d_{i}\leftarrow\frac{\left( {r'_{j}}{r'_{k}}\sin ({\alpha'_{j}}-{\alpha'_{k}}) \right)}{{\left( {r'_{k}}\sin ({\theta_{i}}-{\alpha'_{k}})+{r'_{j}}\sin ({\alpha'_{j}}-{\theta_{i}}) \right)}}$;\\
    }
Return $\boldsymbol{e}\leftarrow\left( d_{1},d_{2},\ldots ,d_{N}\right)$.
\end{algorithm}

Because of the central symmetry of the OBB, the encoded parameters $\mathbf{e}\text{ = (}{{\text{d}}_{1}}\text{,}{{\text{d}}_{2}}\text{,}\cdots \text{,}{{\text{d}}_{N}}\text{)}$ can actually represent $2N$ boundary points distributed around the OBB. In addition, due to the central symmetry, the process of obtaining the OBB parameters is equivalent to sampling from a periodic function with period $\pi$ in the interval $[0,\pi )$ . And the rotation of the OBB is equivalent to the translation of the periodic function. The periodicity ensures the natural continuity in the boundary cases, which is helpful for improving the performance of the network. The detailed discussion can be referred to Section \ref{discussion} of the paper.

\subsection{Polar Decoding}
\begin{figure*}[ht]
\centering
\includegraphics[scale=0.9]{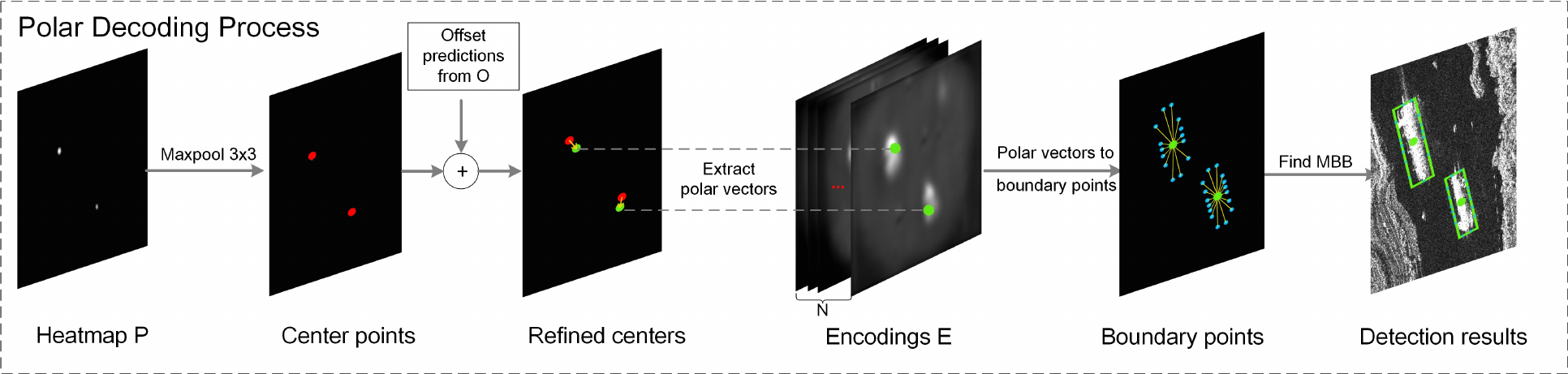}
\caption{The diagram of the polar decoding process, where the outputs of the detection branches are combined to produce the detection results. The center heatmap P and the center offset map O are used for obtaining the center locations of the ship targets. The detection results can be achieved through boundary point extraction and finding the minimum bounding boxes(MBB) from the predicted encoding map E.}
\label{fig7}
\end{figure*}

The overall diagram of the polar decoding process is shown in Fig. \ref{fig7}, and the processing steps are given in Algorithm 2. The polar decoding process decodes the information from the center heatmap $P\in {{\mathbb{R}}^{\tilde{H}\times \tilde{W}\times 1}}$, the center offset map $O\in {{\mathbb{R}}^{\tilde{H}\times \tilde{W}\times 2}}$ and the encoding map $E\in {{\mathbb{R}}^{\tilde{H}\times \tilde{W}\times N}}$ into the detection results. To be specific, firstly a 3×3 Maxpooling layer is employed to process the center heatmap $P$ and output $P’$. The ship centers $C:\{({{x}_{j}},{{y}_{j}})\left| j=1,2,\ldots K \right.\}$ are collected by finding points such that $Maxpoo{l}_{3x3}(P) = P $, where $K$ denotes the number of the detected centers. For each center point $({{x}_{i}},{{y}_{i}})$, the predicted downsampling quantization errors $(\Delta {{x}_{i}},\Delta {{y}_{i}})$ can be obtained from the corresponding location of the offset map $O$. Hence, the coordinate of the $i_{th}$ refined center point can be represented as $({{x}_{i}}+\Delta {{x}_{i}},{{y}_{i}}+\Delta {{y}_{i}})$.

Next, for each detected ship center, the $N$ channels of values from the corresponding location in $E$ are extracted as the predicted OBB parameters. Let $\mathbf{e}=({{d}_{1}},{{d}_{2}},\cdots ,{{d}_{N}})$ denote the predicted OBB parameters of the $i_{th}$ ship target . Since the parameters represent the distances between the boundary points and the center point of the OBB at fixed angles, the boundary point set $S$ of the OBB can be restored as:
\begin{IEEEeqnarray}{l}\label{equ7}
    \begin{aligned}
    S=&\left\{ \left( j\alpha ,{{d}_{j}} \right)\left| j= \right.0,1,\ldots N-1 \right\} \cup \\
    &\left\{ \left( \pi +j\alpha ,{{d}_{j}} \right)\left| j= \right.0,1,\ldots N-1 \right\}
    \end{aligned}
\end{IEEEeqnarray}
where $\alpha {=\pi }/{N}$ denotes the sampling interval angle.

\begin{algorithm}[ht]\label{alg_2}
\setstretch{1}          
\caption{Polar Decoding}
\KwIn{The center heatmap $P$, the center offset map $O$, the encoding map $E$. }
\KwOut{The minimum bounding boxes(MBB) of the detected ship targets.}
Find all points satisfying $Maxpool_{3x3}(P) = P$ as the detected ship centers: $C\leftarrow\{({{x}_{i}},{{y}_{i}})\left| i=1,2,\ldots K \right.\}$; \\
Extract the offset values $\{(\Delta {{x}_{i}},\Delta {{y}_{i}})\left| i=1,2,\ldots K \right.\}$ from O, calculate the refined ship centers: $C'\leftarrow\left\{ \left( {{x}_{i}}+\Delta {{x}_{i}},{{y}_{i}}+\Delta {{y}_{i}} \right)\left| i=1,2,\ldots K \right. \right\}$;\\
\For{$i$ $\leftarrow$ 1 to K} { 
    Extract the predicted OBB parameters $\boldsymbol{e}$ from $E$;\\ 
    Transform $\boldsymbol{e}$ to a boundary point set : \quad \quad \quad \quad $S\leftarrow$ \parbox[t]{0.1\linewidth}{$\left\{ \left( j\alpha ,{{d}_{j}} \right)\left| j= \right.0,1,\ldots N-1 \right\}\cup $ \\
    $\left\{\left( \pi +j\alpha ,{{d}_{j}} \right)\left| j= \right.0,1,\ldots N-1 \right\}$; } \\
    H $\leftarrow$ ConvexHull(S); \\
    $s_{max}$ $\leftarrow$ +Inf.;\\
    \ForEach { $edge$ ${{E}_{k}}$: ($ x_k,y_k $) $\Leftrightarrow$ ($ x_{k+1},y_{k+1} $) of H:} { 
   Calculate $\left\{ \max _p,\min _p,\max _o,\min _o \right\}$ as Equ.\ref{equ9}; \\
    Calculate two side lengths of $B{{B}_{k}}$: $l_p\leftarrow\max _p-\min _p$,$l_o\leftarrow\max _o-\min _o$; \\
    \If {${{l}_{p}}{{l}_{o}}<$$s_{max}$} {
    $s_{max}\leftarrow{{l}_{p}}{{l}_{o}}$, $\text{MB}{{\text{B}}_{i}}\leftarrow\text{B}{{\text{B}}_{k}}$;}
    }
}
Return all MBBs of the detected ship targets.
\end{algorithm}

The next step is to calculate the minimum bounding box(MBB) of $S$. First, we calculate the convex hull of $S$, denoted by $H=\{(x_{j}^{h},y_{j}^{h})|j=1,2,\ldots M\}$. Next, for the $k_{th}$ edge ${{\overset{\scriptscriptstyle\rightharpoonup}{E}}_{k}}\text{=}\left( x_{k+1}^{h}-x_{k}^{h},y_{k+1}^{h}-y_{k}^{h} \right)$ of $H$, we obtain the unit vectors of its parallel and orthogonal directions by: 
\begin{IEEEeqnarray}{l}\label{equ8}
    \begin{aligned}
        &{{\overset{\scriptscriptstyle\rightharpoonup}{e}}_{kp}}=(x_{k+1}^{h}-x_{k}^{h},y_{k+1}^{h}-y_{k}^{h})/\left| {{{\overset{\scriptscriptstyle\rightharpoonup}{E}}}_{k}} \right|\\
        &{{\overset{\scriptscriptstyle\rightharpoonup}{e}}_{\text{ko}}}=(y_{k+1}^{h}-y_{k}^{h},x_{k}^{h}-x_{k+1}^{h})/\left| {{{\overset{\scriptscriptstyle\rightharpoonup}{E}}}_{k}} \right|
    \end{aligned}
\end{IEEEeqnarray}
where ${{\overset{\scriptscriptstyle\rightharpoonup}{e}}_{kp}}$ and ${{\overset{\scriptscriptstyle\rightharpoonup}{e}}_{ko}}$ denote the unit vectors parallel and orthogonal to the edge, respectively. Then the maximum and minimum projections of the vertexes of $H$ in the parallel and orthogonal directions are calculated by:
\begin{IEEEeqnarray}{l}\label{equ9}
    \begin{aligned}
      & ma{{x}_{p}}=\underset{j}{\mathop{\max }}\,(\left| (x_{j}^{h},y_{j}^{h})\centerdot {{{\overset{\scriptscriptstyle\rightharpoonup}{e}}}_{kp}} \right|)\\
      & mi{{n}_{p}}=\underset{j}{\mathop{\min }}\,(\left| (x_{j}^{h},y_{j}^{h})\centerdot {{{\overset{\scriptscriptstyle\rightharpoonup}{e}}}_{kp}} \right|) \\ 
     & ma{{x}_{o}}=\underset{j}{\mathop{\max }}\,(\left| (x_{j}^{h},y_{j}^{h})\centerdot {{{\overset{\scriptscriptstyle\rightharpoonup}{e}}}_{ko}} \right|)\\
     & mi{{n}_{o}}=\underset{j}{\mathop{\min }}\,(\left| (x_{j}^{h},y_{j}^{h})\centerdot {{{\overset{\scriptscriptstyle\rightharpoonup}{e}}}_{ko}} \right|)
    \end{aligned}
\end{IEEEeqnarray}
where ${{\max }_{p}},{{\min }_{p}}$ stand for the maximum and minimum projection parallel to ${{\overset{\scriptscriptstyle\rightharpoonup}{E}}_{k}}$, and  ${{\max }_{o}},{{\min }_{o}}$ denote the maximum and minimum projection orthogonal to ${{\overset{\scriptscriptstyle\rightharpoonup}{E}}_{k}}$. By calculating the difference between the maximum projection and the minimum projection in two directions, we can estimate the side lengths of the bounding box in the $k_{th}$ direction:
\begin{IEEEeqnarray}{l}\label{equ10}
    {{l}_{p}}={{\max }_{p}}-{{\min }_{p}},{{l}_{o}}={{\max }_{o}}-{{\min }_{o}}
\end{IEEEeqnarray}
where ${{l}_{p}}$ and ${{l}_{o}}$ are the lengths of the two sides of the bounding box, respectively; $area_{k}={{l}_{p}}\cdot {{l}_{o}}$ represents the area of the $k_{th}$ bounding box. For all the edges of $H$, the above calculations are carried out to find the smallest bounding box, which is taken as the estimated MBB of the $i_{th}$ ship target. The process is as follows:
\begin{IEEEeqnarray}{l}\label{equ11}
    MB{{B}_{i}}=\underset{B{{B}_{k}}}{\mathop{\arg \min }}\,(are{{a}_{k}})
\end{IEEEeqnarray}
where $BB_{k}$ is the calculated bounding box in the $k_{th}$ direction. We compute MBBs of all ship targets as the detection results.

\subsection{Loss Function}
The training loss of our method is composed of three parts, corresponding to three OBB detection branches.

For the center prediction branch, we adopt the same training approach as \cite{zhou2019objects}. Firstly, for each ship center in the SAR image, a two-dimensional Gaussian mask is generated at the corresponding position of the ground truth map. The standard deviation $ \sigma $ of the Gaussian distribution is set to $1 / 3$ of the ship width. When two Gaussian masks overlap, the larger value is taken for every overlapped position. The loss for the center prediction branch can be calculated by:
\begin{IEEEeqnarray}{l}\label{equ12}
    {{L}_{\text{h}m}}=-\frac{1}{K}\sum\limits_{xy}{\left\{ 
    \begin{matrix}
       {{{\bar{P}}}_{xy}}^{\ \alpha }\log ({{P}_{xy}}) & if\text{ }{{Y}_{xy}}=1  \\
       {{{\bar{Y}}}_{xy}}^{\ \beta }{{P}^{\ \alpha }_{xy}}\log ({{{\bar{P}}}_{xy}}) & otherwise  \\
    \end{matrix} \right.}
\end{IEEEeqnarray}
where $P\in {{[0,1]}^{\tilde{H}\times \tilde{W}\times 1}}$ denotes the output center heatmap; $Y\in {{[0,1]}^{\tilde{H}\times \tilde{W}\times 1}}$ is the ground truth map; ${{\bar{P}}_{xy}}=1-{{P}_{xy}}$ and ${{\bar{Y}}_{xy}}=1-{{Y}_{xy}}$; $\alpha$ and $\beta$ are hyper parameters to control the attention for difficult samples, which are set empirically to 2 and 4 respectively as in \cite{zhou2019objects}; $K$ is the number of ship targets in the SAR image.

Because the resolution of the output is $1/4$ of the input SAR image. The discrete quantization errors are produced in the downsampling process. The center offset regression branch is used for predicting the errors. The center offset regression branch is supervised with the following loss:
\begin{IEEEeqnarray}{l}\label{equ13}
    {{L}_{off}}=\frac{1}{K}\sum\limits_{k=1}^{N}{\left| {{{\overset{\scriptscriptstyle\frown}{O}}}_{{{{\tilde{c}}}_{k}}}}-(\frac{{{c}_{k}}}{R}-{{{\tilde{c}}}_{k}}) \right|}
\end{IEEEeqnarray}
where $R$ represents the downsampling rate, which is 4 in this paper; ${{c}_{k}}\text{=}(x_{k}^{c},y_{k}^{c})\in {{\mathbb{R}}^{2}}$   denote the coordinates of the $k_{th}$ ship center after downsampling quantization; ${{\tilde{c}}_{k}}=\left\lfloor {{c}_{k}}/4 \right\rfloor $ is the downsampled coordinates of the $k_{th}$ ship target; ${{\overset{\scriptscriptstyle\frown}{O}}_{{{{\tilde{c}}}_{k}}}}$ is the predicted discrete quantization error from the offset map $O$ at corresponding $k_{th}$ center location.

For the encoding regression branch, we use smooth-L1 loss for supervised training. But the encoded boundary points contribute to the detection IOU differently. In order to further guide the training of the encoded parameters directly by the IOU metric, we propose to use the IOU metric to weight the smooth-L1 loss. The final loss for the encoding regression branch is calculated as follows:
\begin{IEEEeqnarray}{l}\label{equ14}
    \begin{aligned}
      & {{L}_{encode}}=\frac{1}{K}\sum\limits_{k=1}^{N}{(1+\gamma \frac{-\log (\text{IOU})}{\left| {{L}_{s}}({{{\hat{E}}}_{{{c}_{k}}}},{{e}_{k}}) \right|}){{L}_{s}}({{{\hat{E}}}_{{{c}_{k}}}},{{e}_{k}})} \\ 
     & {{L}_{s}}({{x}_{1}},{{x}_{2}})=\left\{ \begin{matrix}
       0.5{{({{x}_{1}}-{{x}_{2}})}^{2}} & \left| x \right|<1  \\
       \left| {{x}_{1}}-{{x}_{2}} \right|-0.5 & otherwise  \\
    \end{matrix} \right. \\ 
    \end{aligned}
\end{IEEEeqnarray}
where ${{\hat{E}}_{{{c}_{k}}}}$ and $e_{k}$ stand for the predicted and the actual encoded parameters of the $k_{th}$ ship target; IOU represents the IOU calculation operation for the predicted and the actual OBB of the $k_{th}$ ship target; ${{L}_{s}}(\centerdot )$ is the smooth L1 regression loss; $\gamma$ is the weight parameter, which is set to 1 in our experiment. By dividing the magnitude of the smooth-l1 loss, the IOU metric provides the gradient magnitude and the smooth-l1 loss determines the direction of the gradient. 

Finally, the three parts of the loss are combined to form the overall loss function:
\begin{IEEEeqnarray}{l}\label{equ15}
    {L_{total}}={{L}_{hm}}+{L_{off}}+{L_{encode}}
\end{IEEEeqnarray}

\begin{table*}[!t]
\footnotesize
\renewcommand{\arraystretch}{1.3}
\caption{The detailed information of RSSDD dataset}
\label{tab1}
\centering
\begin{tabular}{p{1.5cm}<{\centering}p{2.5cm}<{\centering}p{2.5cm}<{\centering}p{1.5cm}<{\centering}p{2.5cm}<{\centering}p{2.5cm}<{\centering}}
\toprule
Satellite & Polarization & Resolution & Waveband & Imaging Mode & Location \\
\midrule
\makecell[c]{Sentinel-1 \\ RadarSat-2 \\ TerraSAR-X} & \makecell[c]{HH,VV,HV,VH} & \makecell[c]{1m-15m} & \makecell[c]{C,X} & \makecell[c]{Interferometric, \\ wide swath Mode, \\ Spotlight Mode, \\ Strip Mode}
 & \makecell[c]{Yantai, China, \\ Visakhapatnam, India} \\
\bottomrule
\end{tabular}
\end{table*}

\subsection{Dataset Description and Experimental Settings}

\begin{figure}[t]
\centering
\subfloat[]{
\begin{minipage}[t]{2.7cm}
\centering
\includegraphics[width=1\linewidth,height=0.8\linewidth]{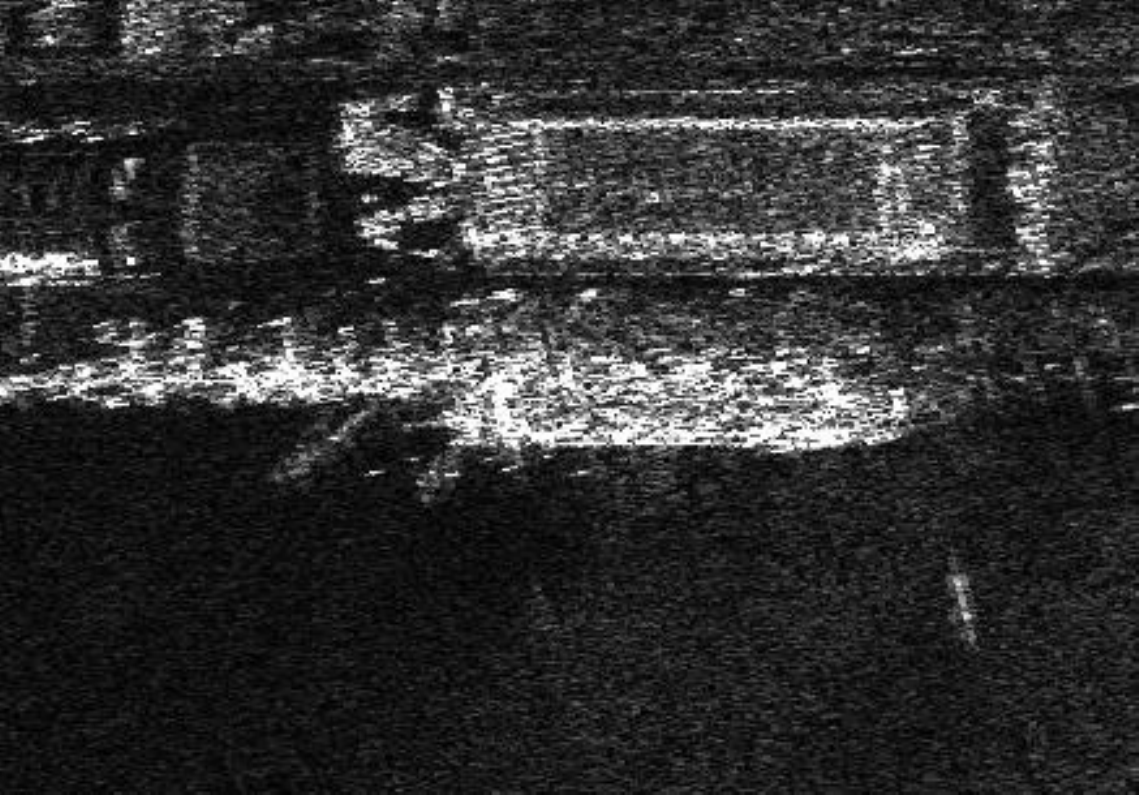}\vspace{2pt}
\end{minipage}
}\hspace{-8pt}
\subfloat[]{
\begin{minipage}[t]{2.7cm}
\centering
\includegraphics[width=1\linewidth,height=0.8\linewidth]{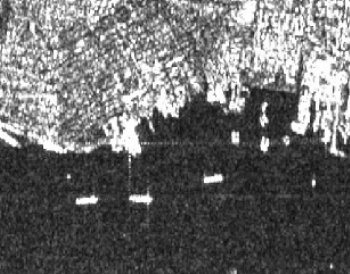}\vspace{2pt}
\end{minipage}
}\hspace{-8pt}
\subfloat[]{
\begin{minipage}[t]{2.7cm}
\centering
\includegraphics[width=1\linewidth,height=0.8\linewidth]{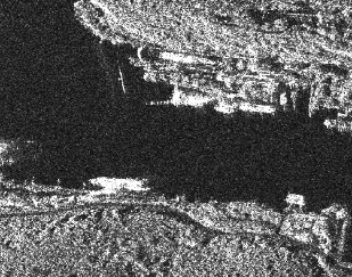}\vspace{2pt}
\end{minipage}
}\hspace{-8pt}
\subfloat[]{
\begin{minipage}[t]{2.7cm}
\centering
\includegraphics[width=1\linewidth,height=0.8\linewidth]{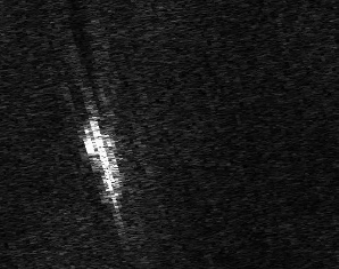}\vspace{2pt}
\end{minipage}
}\hspace{-8pt}
\subfloat[]{
\begin{minipage}[t]{2.7cm}
\centering
\includegraphics[width=1\linewidth,height=0.8\linewidth]{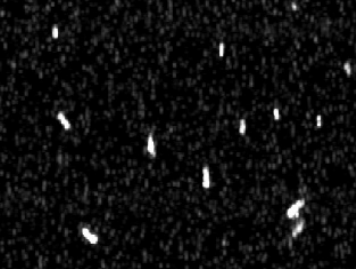}\vspace{2pt}
\end{minipage}
}\hspace{-8pt}
\subfloat[]{
\begin{minipage}[t]{2.7cm}
\centering
\includegraphics[width=1\linewidth,height=0.8\linewidth]{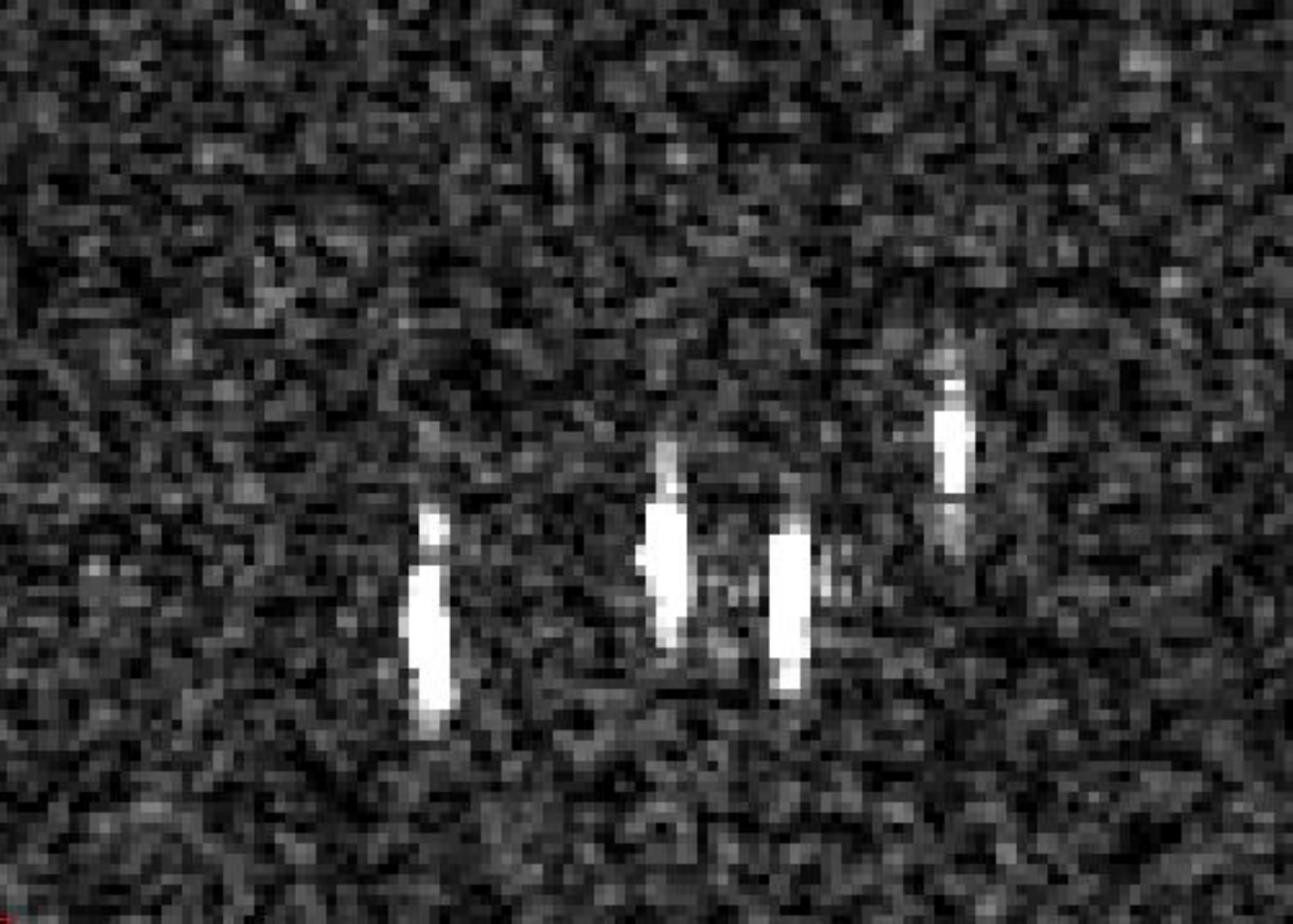}\vspace{2pt}
\end{minipage}
}

\caption{Several SAR images in the RSSDD dataset.}
\label{fig8}
\end{figure}

\section{Experimental Results}
In this section, we report the experiments carried out on the RSSDD dataset in detail to evaluate the effectiveness of our proposed method. Firstly, the information of the dataset and the experimental settings used in this paper are described. Then, the evaluation metrics are illustrated. Next, qualitative and quantitative comparison results between the proposed method and other OBB encoding schemes and other detection methods are given to verify the effectiveness of the proposed method.

RSSDD is a publicly available OBB-based SAR ship detection dataset \cite{li2017ship}, composed of SAR images of multi resolution, multi polarization and multi scene. The specific information of RSSDD is given in Table \ref{tab1}. RSSDD contains 1160 SAR images and 2456 ship targets, all of which are labeled with four corners of the OBB. The sizes of the SAR images are different from one another, from the smallest $217\times214$ to the largest $526\times646$. The average size of the images is $481\times331$. In the experiments, the dataset is divided into the training set and the test set with the ratio of 8:2. To be specific, the training set contains 928 SAR images and the test set contains 232 SAR images. The dataset contains a variety of scenes. As shown in Fig. \ref{fig8}, Fig. \ref{fig8}(a)-(c) give the examples of images with inshore scenes, Fig. \ref{fig8}(d)-(f) show several offshore scene images. As can be seen, Fig. \ref{fig8}(a),(d) contain large ship targets, while other images exhibit small scale ship targets. Compared with the offshore scenes, the inshore scenes contain more land clutters, which make the detection more difficult. Besides, the number of inshore scene images is less. In order to better evaluate the performance of the detector in different scenes, we further split the test set into two kinds of scenes: the inshore scenes and the offshore scenes, which contain 39 and 193 SAR images, respectively.

The input SAR images are resized to $608\times608$ in both the training and the inference stage, and the output feature $F_{final}$ is with the resolution of $152\times152$. In the training stage, we use the ImageNet pre-trained weights to initialize the parameters of the feature extraction backbone. The hyper parameter N in the polar encoding process is set to 8. Adaptive moment estimation (Adam) optimizer \cite{kingma2014adam} is adopted as the training optimizer, the weight decay of which is 0.0005. The initial learning rate is set to $1.25\times {{10}^{\text{-}4}}$. The learning rate is then adjusted according to the exponential decay rule. The mini-batch size used in the stochastic gradient descent algorithm is 8. The model is trained in a total of 150 epochs. The algorithm is implemented with the deep learning framework Pytorch \cite{paszke2019pytorch}. The comparison experiments are conducted based on the framework proposed by \cite{chen2019mmdetection}. All the experiments are carried out on the platform with Ubuntu18.04 system, 32G memory and Tesla P100 GPU .

\subsection{Evaluation Metrics}
Three widely adopted metrics, the precision-recall curve(PR curve), AP and F1 are used to evaluate the performance of the models. For the PR curve, the recall rate ($R_{d}$) is taken as the x-axis and the precision rate ($P_{d}$) is taken as the y-axis, which can be calculated as follows:
\begin{IEEEeqnarray}{l}\label{equ16}
    \begin{aligned}
      & {{P}_{d}}=\frac{{{N}_{td}}}{{{N}_{d}}} \\ 
     & {{R}_{d}}=\frac{{{N}_{td}}}{{{N}_{r}}} \\ 
    \end{aligned}
\end{IEEEeqnarray}
where $N_{td}$ is the number of the correctly detected targets, $N_{d}$ denotes the total number of the detected targets, and $N_{r}$ represents the actual number of the targets. The AP metric quantitatively evaluates the comprehensive detection performance of the detector by calculating the area under the PR curve as follows:
\begin{IEEEeqnarray}{l}\label{equ17}
    \text{AP}=\int_{0}^{1}{{{\text{P}}_{d}}\text{(}{{\text{R}}_{d}}\text{)}d{{\text{R}}_{d}}}
\end{IEEEeqnarray}

AP measures the overall detection performance of the detector under different thresholds. And the F1 metric indicates the comprehensive performance of the detector under the single-point threshold. As F1 varies with thresholds, we take the maximum F1 under all thresholds for comparison. The F1 metric is defined as:
\begin{IEEEeqnarray}{l}\label{equ18}
    {{\text{F}}_{1}}\text{=}2\cdot \frac{{{P}_{d}}\cdot {{R}_{d}}}{{{P}_{d}}+{{R}_{d}}}
\end{IEEEeqnarray}

\subsection{Comparison with different OBB encoding schemes}
In order to evaluate the effectiveness of the polar encoding scheme, we implement three different OBB encoding schemes based on the same center-point-based detection framework \cite{zhou2019objects}, including the angle-based, the point-based and the proposed polar-based encoding scheme. These encoding schemes all adopt the center prediction branch and the offset regression branch to locate the center point of the target. But they represent the OBB of the ship targets in different ways. Among them, the angle-based scheme is the 90°-based representation introduced in Section \ref{bdp}. It represents the position and shape of the OBB by the center point, width, height and the rotation angle of the OBB. The detailed introduction of point-based encoding scheme can be referred to \cite{yi2021oriented}. This method represents OBB by the center point and four vectors pointing from the center point to the midpoint of four edges of the OBB. In order to reduce the boundary discontinuity problem, the method also distinguishes HBB from OBB by training a classification branch. Different regression rules are applied in the HBB and OBB prediction process.

\begin{figure}[t]
\centering  
\subfloat[]{  
\includegraphics[scale=0.3]{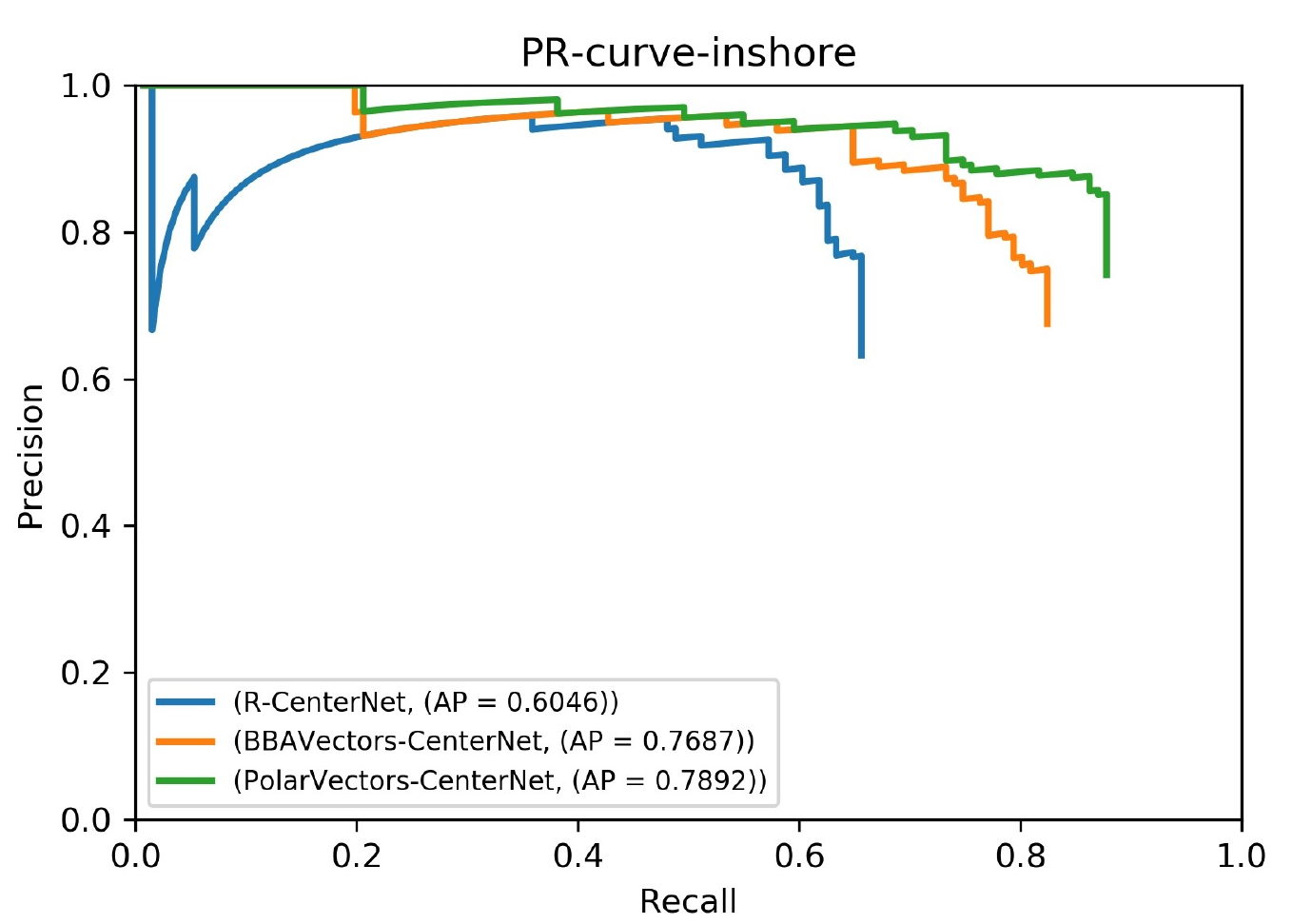}}
\hspace{2pt} 
\subfloat[]{  
\includegraphics[scale=0.3]{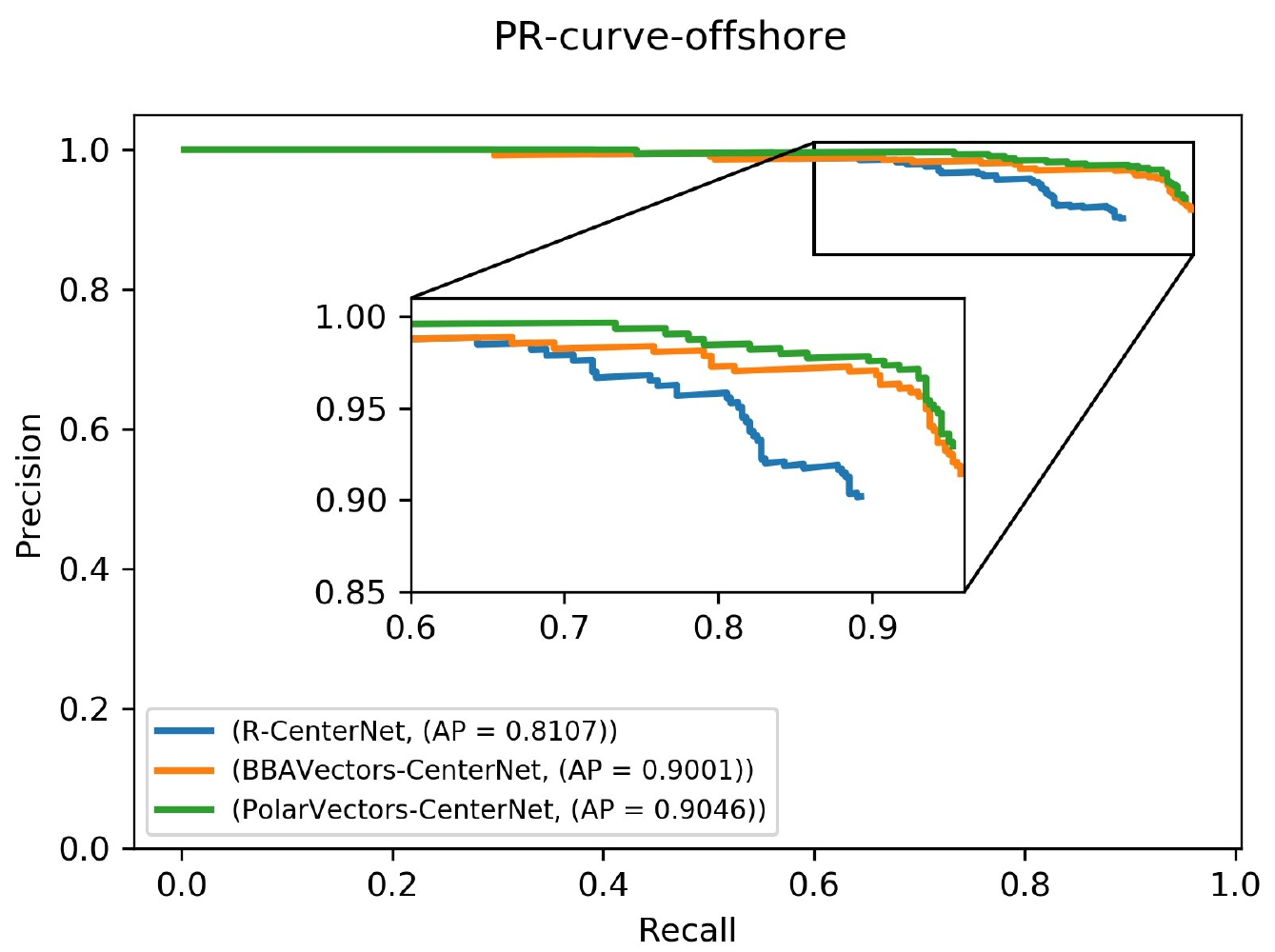}}  
\caption{PR curves of different OBB encoding schemes. (a) PR curves for the inshore scenes; (b) PR curves for the offshore scenes.}  
\label{fig9}
\end{figure}

\begin{figure}[t]
\centering
\subfloat[]{
\begin{minipage}[t]{2.05cm} 
\centering
\includegraphics[width=1\linewidth,height=0.9\linewidth]{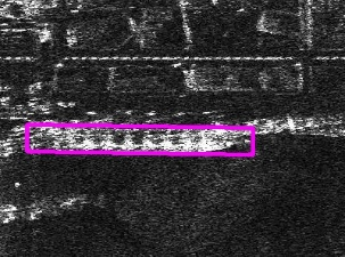}\vspace{2pt}
\includegraphics[width=1\linewidth,height=0.9\linewidth]{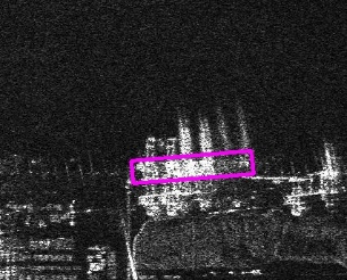}\vspace{2pt}
\includegraphics[width=1\linewidth,height=0.9\linewidth]{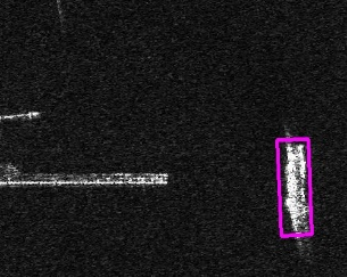}
\end{minipage}
}\hspace{-8pt} 
\subfloat[]{
\begin{minipage}[t]{2.05cm}
\centering
\includegraphics[width=1\linewidth,height=0.9\linewidth]{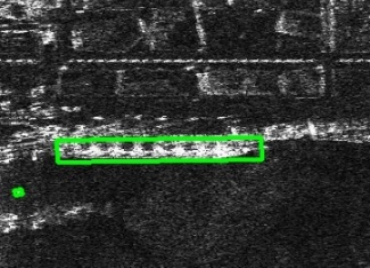}\vspace{2pt}
\includegraphics[width=1\linewidth,height=0.9\linewidth]{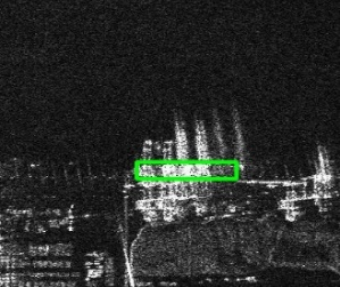}\vspace{2pt}
\includegraphics[width=1\linewidth,height=0.9\linewidth]{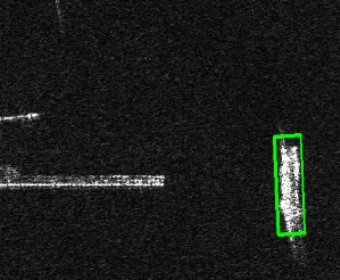}
\end{minipage}
}\hspace{-8pt}
\subfloat[]{
\begin{minipage}[t]{2.05cm}
\centering
\includegraphics[width=1\linewidth,height=0.9\linewidth]{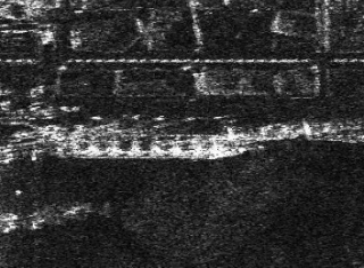}\vspace{2pt}
\includegraphics[width=1\linewidth,height=0.9\linewidth]{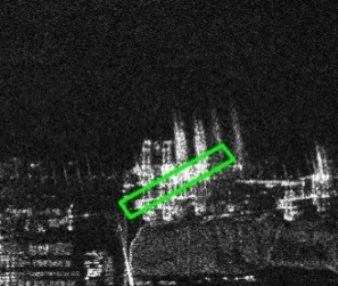}\vspace{2pt}
\includegraphics[width=1\linewidth,height=0.9\linewidth]{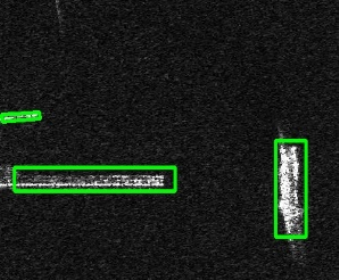}
\end{minipage}
}\hspace{-8pt}
\subfloat[]{
\begin{minipage}[t]{2.05cm}
\centering
\includegraphics[width=1\linewidth,height=0.9\linewidth]{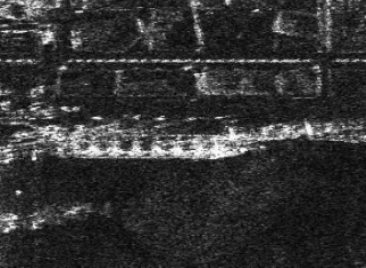}\vspace{2pt}
\includegraphics[width=1\linewidth,height=0.9\linewidth]{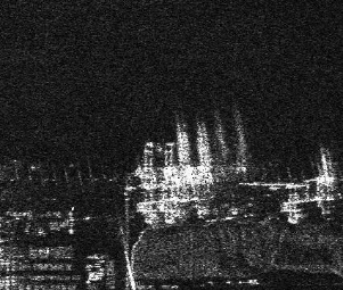}\vspace{2pt}
\includegraphics[width=1\linewidth,height=0.9\linewidth]{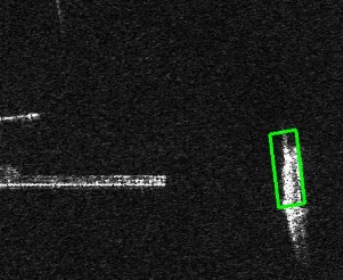}
\end{minipage}
}
\caption{The detection results of different encoding schemes in the boundary cases, the green rectangles denote the detected ship targets. (a) Ground truth; (b) our method; (c) the point-based method; (d) the angle-based method.}
\label{fig10}
\end{figure}

\begin{table*}[t]
\caption{The detection metrics of different OBB encoding schemes.The bold items denote the largest values in the columns.}
\label{tab2}
\centering
\begin{tabular}{c|cccc|cccc}
\toprule
\multirow{2}{*}{Methods} & \multicolumn{4}{c|}{Inshore}      & \multicolumn{4}{c}{Offshore}     \\ \cline{2-9} 
                         & $P_{d}$     & $R_{d}$      & F1      & AP     & $P_{d}$     & $R_{d}$   & F1      & AP\\ \hline
Angle-based         & 0.6324 & 0.6565 & 0.6442 & 0.6046 & 0.9018 & 0.8928 & 0.8972 & 0.8107 \\ 
Point-based         & 0.7879 & 0.7939 & 0.7909 & 0.7687 & 0.9401 & \textbf{0.9401} & 0.9401 & 0.9001 \\ 
Polar-based(Ours)   & \textbf{0.8720} & \textbf{0.8321} & \textbf{0.8594} & \textbf{0.7942} & \textbf{0.9519} & 0.9377 & \textbf{0.9447} & \textbf{0.9046} \\ 
\bottomrule
\end{tabular}
\end{table*}

In order to quantitatively measure the detection performance of the three encoding schemes, the detection metrics of the three encoding schemes are listed in Table \ref{tab2}. It can be seen that the F1 and AP metric of our method are higher than those of the other two methods in both scenes. In addition, the PR curves of the three encoding schemes are shown in Fig. \ref{fig9}. We can see that the PR curve of our method lies outer than those of the other two methods, indicating that our method has better detection performance. Because of the boundary discontinuity problem and the IOU sensitivity problem in the angle regression, the detection performance of the angle-based method is worse than the other two methods in both scenes. To overcome the boundary discontinuity problem, an extra classification branch is required for the point-based method to distinguishing HBB from OBB. It leads to extra training objectives and the decline of the detection performance. For our method, the boundary discontinuity problem is addressed by the specially designed polar encoding and decoding process. The training objectives are more direct and more concentrated, thus the overall detection performance is improved.

\begin{table*}[t]
\caption{The detection results of different methods. The bold items denote the optimal values in the columns, \\ the underlined items represent the suboptimal values in the columns.}
\label{tab3}
\centering
\begin{tabular}{c|cccc|cccc|c}
\toprule
\multirow{2}{*}{Methods} & \multicolumn{4}{c|}{Inshore}      & \multicolumn{4}{c|}{Offshore}     \\ \cline{2-10} 
                         & $P_{d}$     & $R_{d}$     & F1     & AP     & $P_{d}$     & $R_{d}$     & F1     & AP    & Time(ms)\\ \hline
Ours(w IOU Loss)         & \textbf{0.8800} & \textbf{0.8397} & \textbf{0.8594} & \textbf{0.7942} & {\ul 0.9519} & 0.9377 & \textbf{0.9447} & \textbf{0.9046} & {\ul 45}\\ 
Ours(w/o   IOU Loss)     & {\ul 0.8720} & {\ul 0.8321} & {\ul 0.8516} & {\ul 0.7899} & 0.9383 & \textbf{0.9476} & 0.9429 & 0.8992 & {\ul 45} \\ 
BBAVectors               & 0.7879 & 0.7933 & 0.7909 & 0.7687 & 0.9401 & {\ul 0.9401} & 0.9401 & 0.9001 & 50 \\ 
ROITransformer           & 0.7027 & 0.7939 & 0.7455 & 0.7540 & \textbf{0.9542} & 0.9352 & {\ul 0.9446} & {\ul 0.9026} & 54 \\ 
FRCNN-OBB                & 0.6266 & 0.7557 & 0.6851 & 0.7064 & 0.9086 & 0.8928 & 0.9006 & 0.8749 & 53 \\ 
RetinaNet-OBB            & 0.4286 & 0.4809 & 0.4532 & 0.4051 & 0.8005 & 0.8803 & 0.8385 & 0.7830 & \textbf{41} \\ \bottomrule
\end{tabular}
\end{table*}

Fig. \ref{fig10} gives the detection results of different encoding schemes in the boundary cases. Fig. \ref{fig10}(a) gives the ground truth and Fig. \ref{fig10}(b)-(d) show the detection results of our method, the point-based method and the angle-based method. It can be seen from Fig. \ref{fig10}(b) that the proposed method can accurately locate the ship targets in the horizontal direction, which benefits from the boundary continuity of our polar encoding scheme. As can be seen from Fig. \ref{fig10}(c), the point-based encoding scheme fails to detect the ship target in the first SAR image. While in the second image, the orientation of the ship target is mispredicted. For the third row, the ship target and the land clutter are both located by HBB. It is because that the point-based encoding scheme has to introduce the extra network branch and loss to distinguish HBB from OBB, which increases the difficulty in network training. The results of the angle-based method shown in Fig. \ref{fig10}(d) indicate missed detections for inshore scenes. And the result in the offshore scene is inaccurate. This is due to the boundary discontinuity and the IOU sensitivity in the angle regression.

\subsection{Comparison with other OBB-based ship detection methods}
In this section, we compare our method with several state-of-the-art OBB-based ship detectors, including Box Boundary Aware Vectors (BBAVectors) \cite{yi2021oriented}, Region of Interest Transformer (ROITransformer) \cite{ding2019learning}, OBB-based Faster-RCNN (FRCNN-OBB) \cite{xia2018dota} and OBB-based RetinaNet (RetinaNet-OBB). BBAVectors is an anchor-free detection method, which combines CenterNet \cite{zhou2019objects} with point-based encoding scheme. ROITransformer is an anchor-based method, which transforms the horizontal ROI into rotating ROI through training, so as to improve the feature extraction ability for arbitrary-oriented targets. FRCNN-OBB is a two-stage detection method based on Faster-RCNN. It first extracts features of the candidate target regions, which are then used to detect targets by predicting the errors between anchors and the OBBs. RetinaNet-OBB is a one-stage method based on RetinaNet, which directly regresses the errors between anchors and OBBs without extracting candidate regions.

\begin{figure*}[t]
\centering
\subfloat[]{
\begin{minipage}[t]{2.8cm} 
\centering
\includegraphics[width=1\linewidth,height=0.8\linewidth]{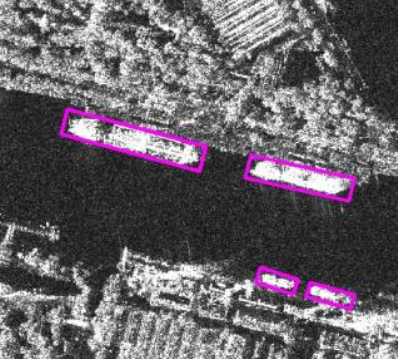}\vspace{2pt}
\includegraphics[width=1\linewidth,height=0.8\linewidth]{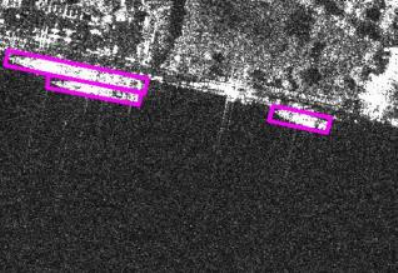}\vspace{2pt}
\includegraphics[width=1\linewidth,height=0.8\linewidth]{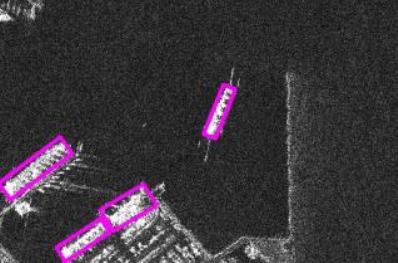}\vspace{2pt}
\includegraphics[width=1\linewidth,height=0.8\linewidth]{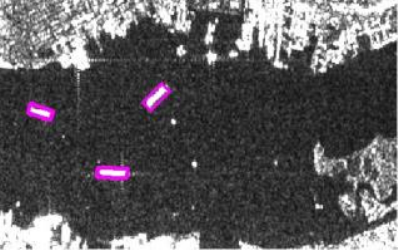}\vspace{2pt}
\includegraphics[width=1\linewidth,height=0.8\linewidth]{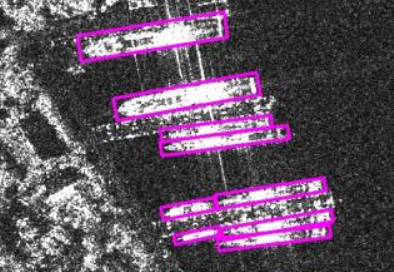}\vspace{2pt}
\includegraphics[width=1\linewidth,height=0.8\linewidth]{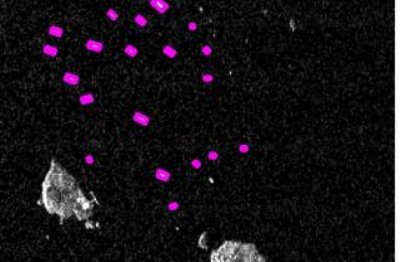}
\end{minipage}
}\hspace{-8pt} 
\subfloat[]{
\begin{minipage}[t]{2.8cm}
\centering
\includegraphics[width=1\linewidth,height=0.8\linewidth]{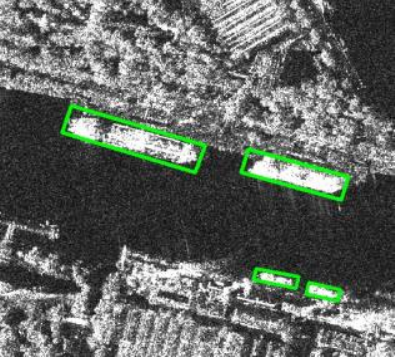}\vspace{2pt}
\includegraphics[width=1\linewidth,height=0.8\linewidth]{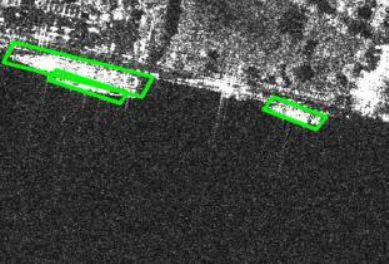}\vspace{2pt}
\includegraphics[width=1\linewidth,height=0.8\linewidth]{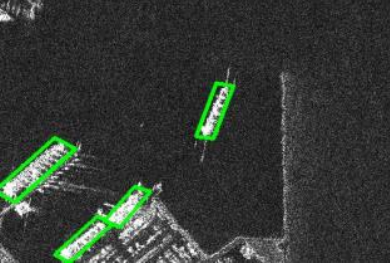}\vspace{2pt}
\includegraphics[width=1\linewidth,height=0.8\linewidth]{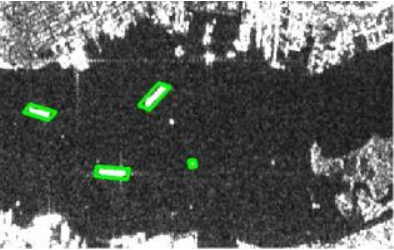}\vspace{2pt}
\includegraphics[width=1\linewidth,height=0.8\linewidth]{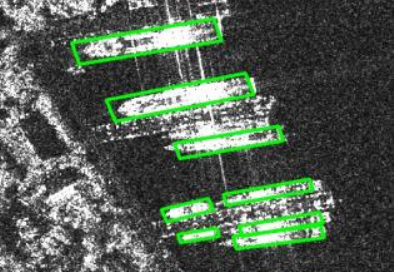}\vspace{2pt}
\includegraphics[width=1\linewidth,height=0.8\linewidth]{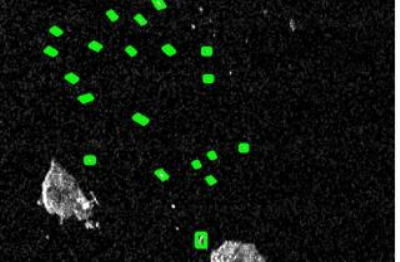}
\end{minipage}
}\hspace{-8pt}
\subfloat[]{
\begin{minipage}[t]{2.8cm}
\centering
\includegraphics[width=1\linewidth,height=0.8\linewidth]{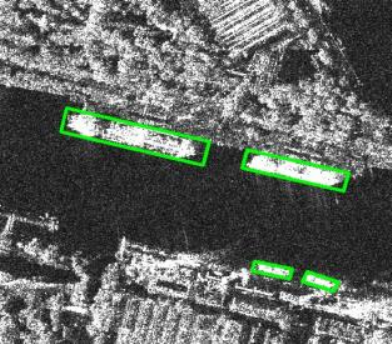}\vspace{2pt}
\includegraphics[width=1\linewidth,height=0.8\linewidth]{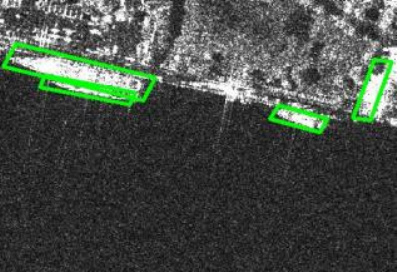}\vspace{2pt}
\includegraphics[width=1\linewidth,height=0.8\linewidth]{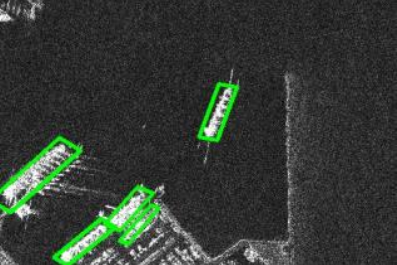}\vspace{2pt}
\includegraphics[width=1\linewidth,height=0.8\linewidth]{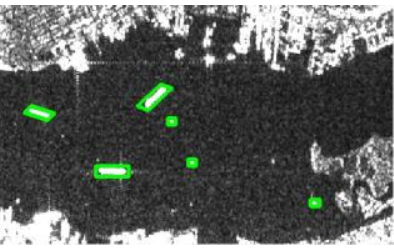}\vspace{2pt}
\includegraphics[width=1\linewidth,height=0.8\linewidth]{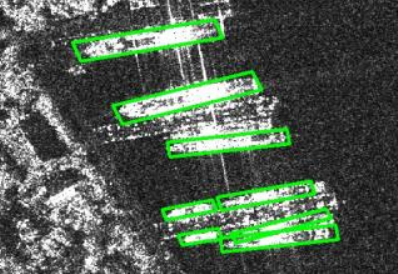}\vspace{2pt}
\includegraphics[width=1\linewidth,height=0.8\linewidth]{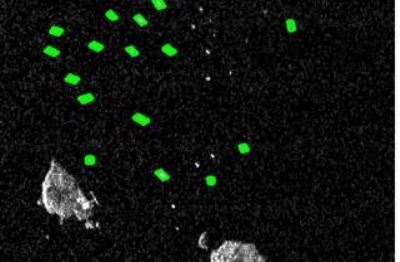}
\end{minipage}
}\hspace{-8pt}
\subfloat[]{
\begin{minipage}[t]{2.8cm}
\centering
\includegraphics[width=1\linewidth,height=0.8\linewidth]{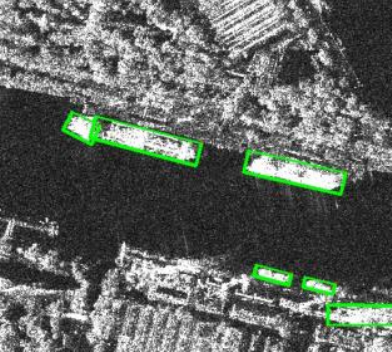}\vspace{2pt}
\includegraphics[width=1\linewidth,height=0.8\linewidth]{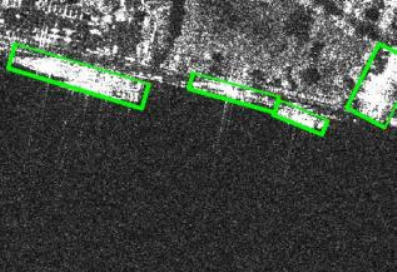}\vspace{2pt}
\includegraphics[width=1\linewidth,height=0.8\linewidth]{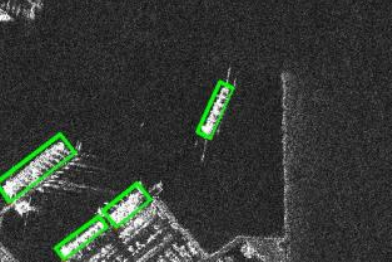}\vspace{2pt}
\includegraphics[width=1\linewidth,height=0.8\linewidth]{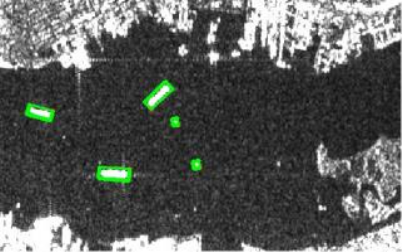}\vspace{2pt}
\includegraphics[width=1\linewidth,height=0.8\linewidth]{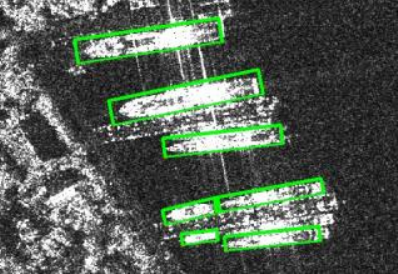}\vspace{2pt}
\includegraphics[width=1\linewidth,height=0.8\linewidth]{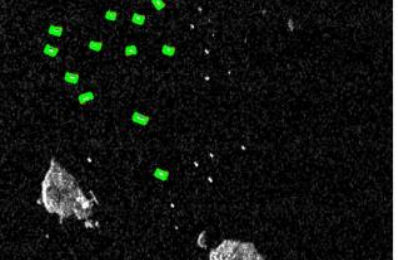}
\end{minipage}
}\hspace{-8pt}
\subfloat[]{
\begin{minipage}[t]{2.8cm}
\centering
\includegraphics[width=1\linewidth,height=0.8\linewidth]{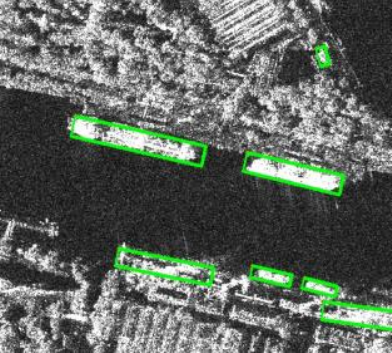}\vspace{2pt}
\includegraphics[width=1\linewidth,height=0.8\linewidth]{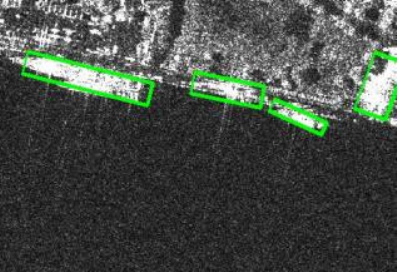}\vspace{2pt}
\includegraphics[width=1\linewidth,height=0.8\linewidth]{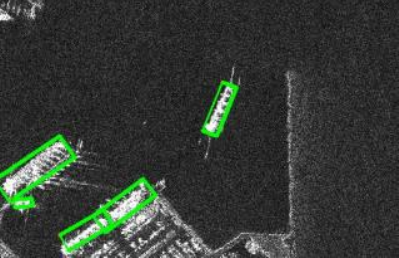}\vspace{2pt}
\includegraphics[width=1\linewidth,height=0.8\linewidth]{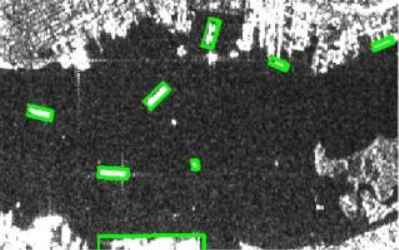}\vspace{2pt}
\includegraphics[width=1\linewidth,height=0.8\linewidth]{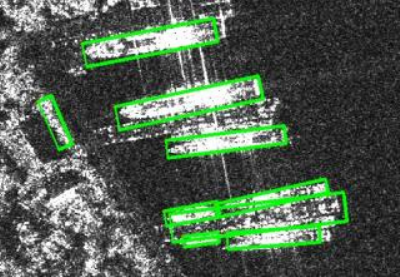}\vspace{2pt}
\includegraphics[width=1\linewidth,height=0.8\linewidth]{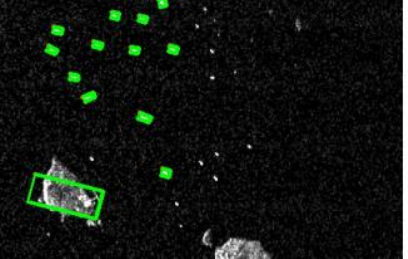}
\end{minipage}
}\hspace{-8pt}
\subfloat[]{
\begin{minipage}[t]{2.8cm}
\centering
\includegraphics[width=1\linewidth,height=0.8\linewidth]{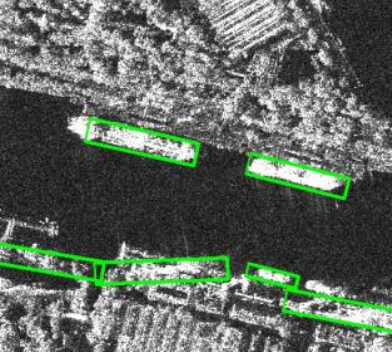}\vspace{2pt}
\includegraphics[width=1\linewidth,height=0.8\linewidth]{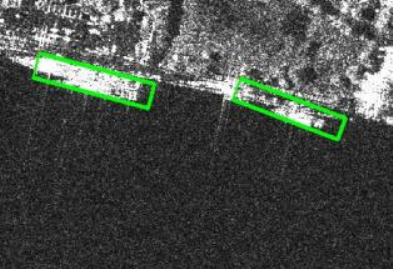}\vspace{2pt}
\includegraphics[width=1\linewidth,height=0.8\linewidth]{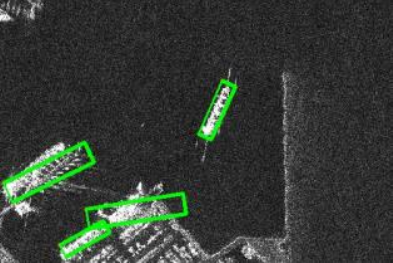}\vspace{2pt}
\includegraphics[width=1\linewidth,height=0.8\linewidth]{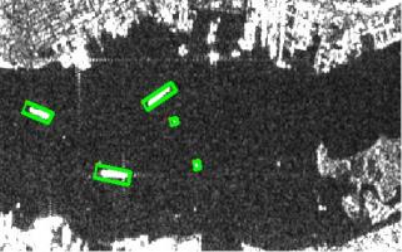}\vspace{2pt}
\includegraphics[width=1\linewidth,height=0.8\linewidth]{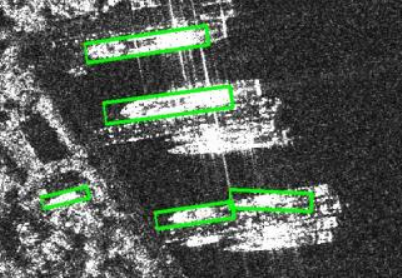}\vspace{2pt}
\includegraphics[width=1\linewidth,height=0.8\linewidth]{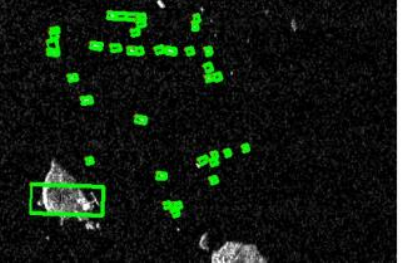}
\end{minipage}
}
\caption{The detection results of different methods. The purple rectangles denote the actual ship targets and the green rectangles are the detection results. (a) Ground truth; (b) Our method; (c) BBAVectors; (d) ROITransformer; (e) FRCNN-OBB; (f) RetinaNet-OBB.}
\label{fig12}
\end{figure*}

Table \ref{tab3} shows the quantitative comparison between our proposed method and other methods on the RSSDD dataset. It can be seen from the table that, without using IOU-weighted regression loss, the F1 and AP metrics of our method are higher than other methods in the inshore scenes, but slightly lower than ROITransformer in the offshore scenes. After using the IOU-weighted regression loss, the F1 and AP of our method are better than other methods in both inshore and offshore scenes. Among other methods, BBAVectors performs relatively better in the inshore scenes, whose F1 metric reaches 0.7909. But there still exists a gap of more than 5\% from our method, and the AP of BBAVecotrs is 2.5\% lower than that of our method. ROITransformer achieves slightly worse performance than BBAVectors in the inshore scenes. And the detection performance of FRCNN-OBB and RetinaNet-OBB is poor compared to other methods in inshore scenes. The above results suggest that our method can effectively avoid the boundary discontinuity problem and improve the detection performance in the complex inshore scenes. For offshore scenes, our proposed method achieves substantial advantages over FRCNN-OBB and RetinaNet-OBB. But the gap between the F1 and AP of our method and that of BBAVectors and ROITransformer is modest. The reason why the gap is not obvious is that the offshore scenes are generally simpler and contain less clutters. As a result, it is easier for the detectors to obtain comparable performance in the offshore scenes, which narrows the gaps between different methods. In summary, we can see from the results that the proposed method can achieve better detection results compared with other methods because it avoids the boundary discontinuity problem and uses IOU-weighted loss to further guide the network training.

To discuss the detection efficiency of different detection methods, the average detection time per image on the test set is also given in Table \ref{tab3}. It can be seen from the table that the detection time of different methods is relatively close. Our method is slightly slower than RetinaNet-OBB and faster than the other four methods. This is because (1) our method adopts an anchor-free framework, which avoids complex calculation of IOU between anchors and the target bounding boxes; (2) our method does not need to add extra network branches to deal with the boundary discontinuity problem; (3) although the calculation of the minimum bounding box is carried out on CPU, which cannot take advantage of the parallel computation capacity of GPU, we can first filter out the low confidence targets before calculating the minimum bounding box, so it will not become the calculation bottleneck.

\begin{figure}[t]
\centering  
\subfloat[]{  
\includegraphics[scale=0.295]{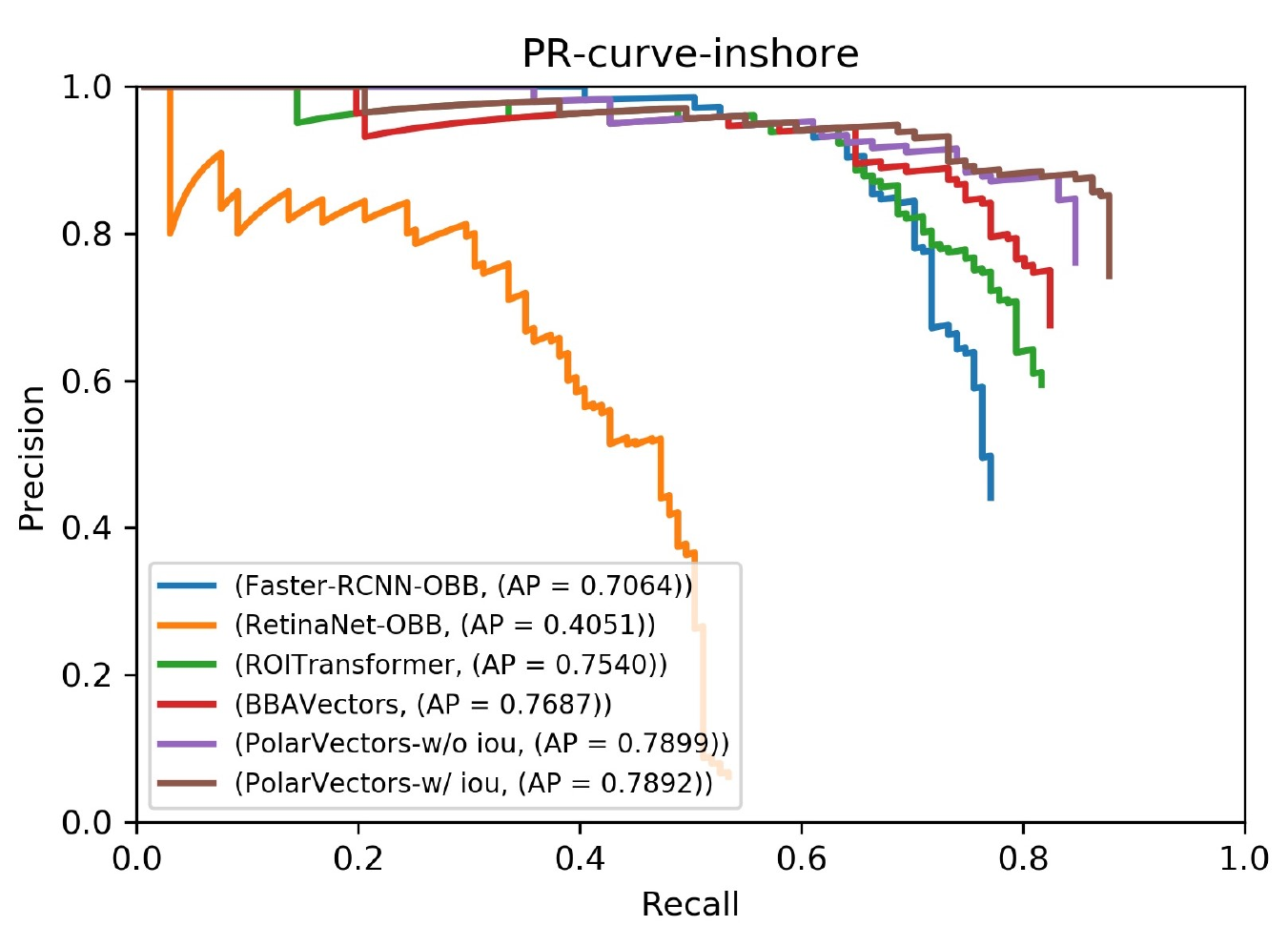}}
\hspace{2pt} 
\subfloat[]{  
\includegraphics[scale=0.305]{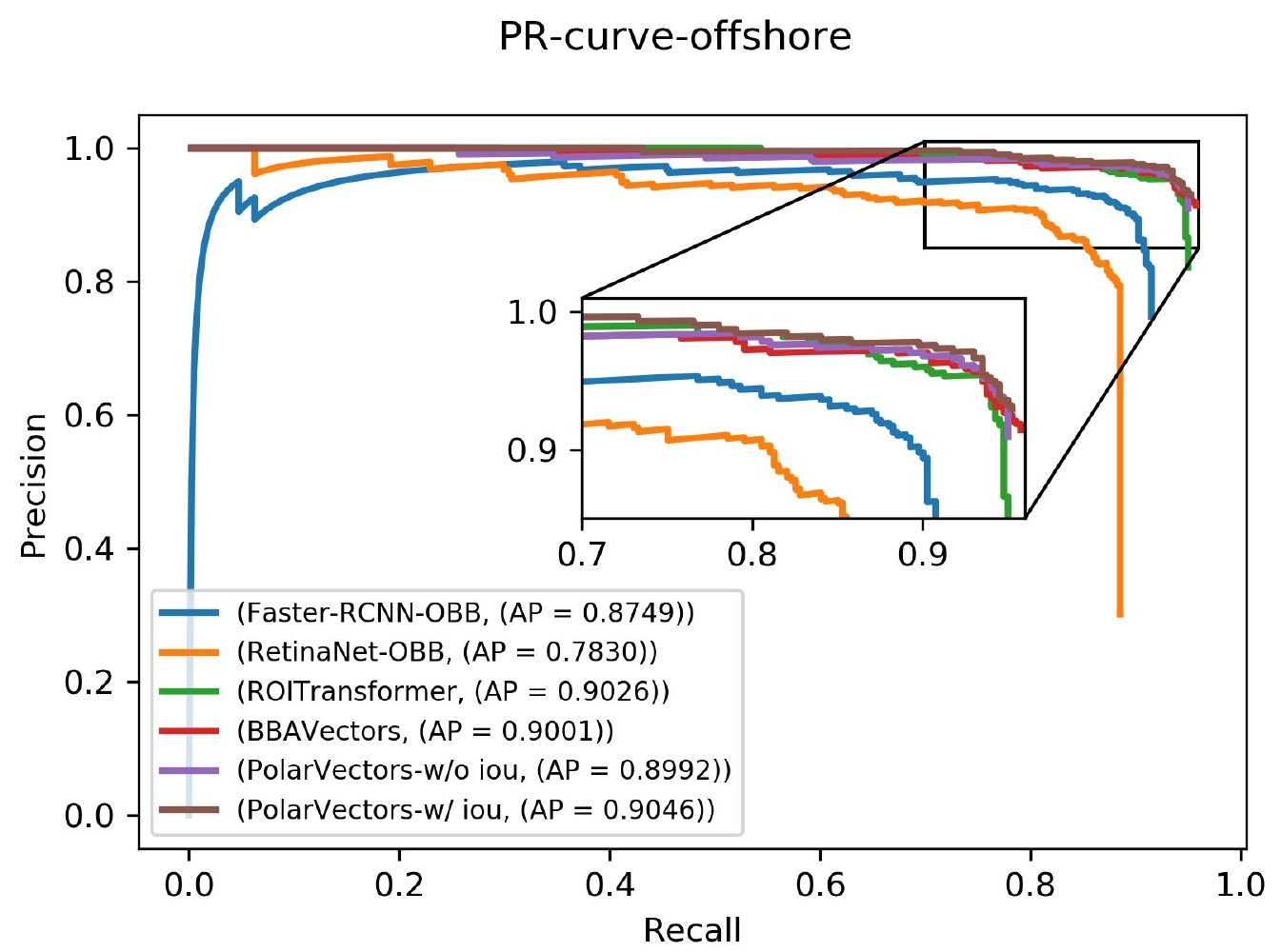}}  
\caption{PR curves of different methods in different scenes. (a) PR curves for inshore scenes;(b) PR curves for offshore scenes.}  
\label{fig11}
\end{figure}

Fig. \ref{fig11} shows the comparison of PR curves of different methods in different scenes. It can be observed from Fig. \ref{fig11}(a) that the PR curve of our method lies outer than other methods regardless of whether the IOU-weighted loss function is used or not, which proves the effectiveness of our method for detecting inshore ships. The PR curve shows improvement after using the IOU-weighted regression loss, which indicates that the IOU-weighted loss can further guide training and improve the detection performance. For the offshore scenes, the PR curves of FRCNN-OBB and RetinaNet-OBB from Fig. \ref{fig11}(b) obviously lie lower than others, showing their poor detection performance. Among other methods, our method lies outer than other methods, demonstrating the effectiveness of our method.

\begin{figure}[!t]
\centering
\includegraphics[scale=0.7]{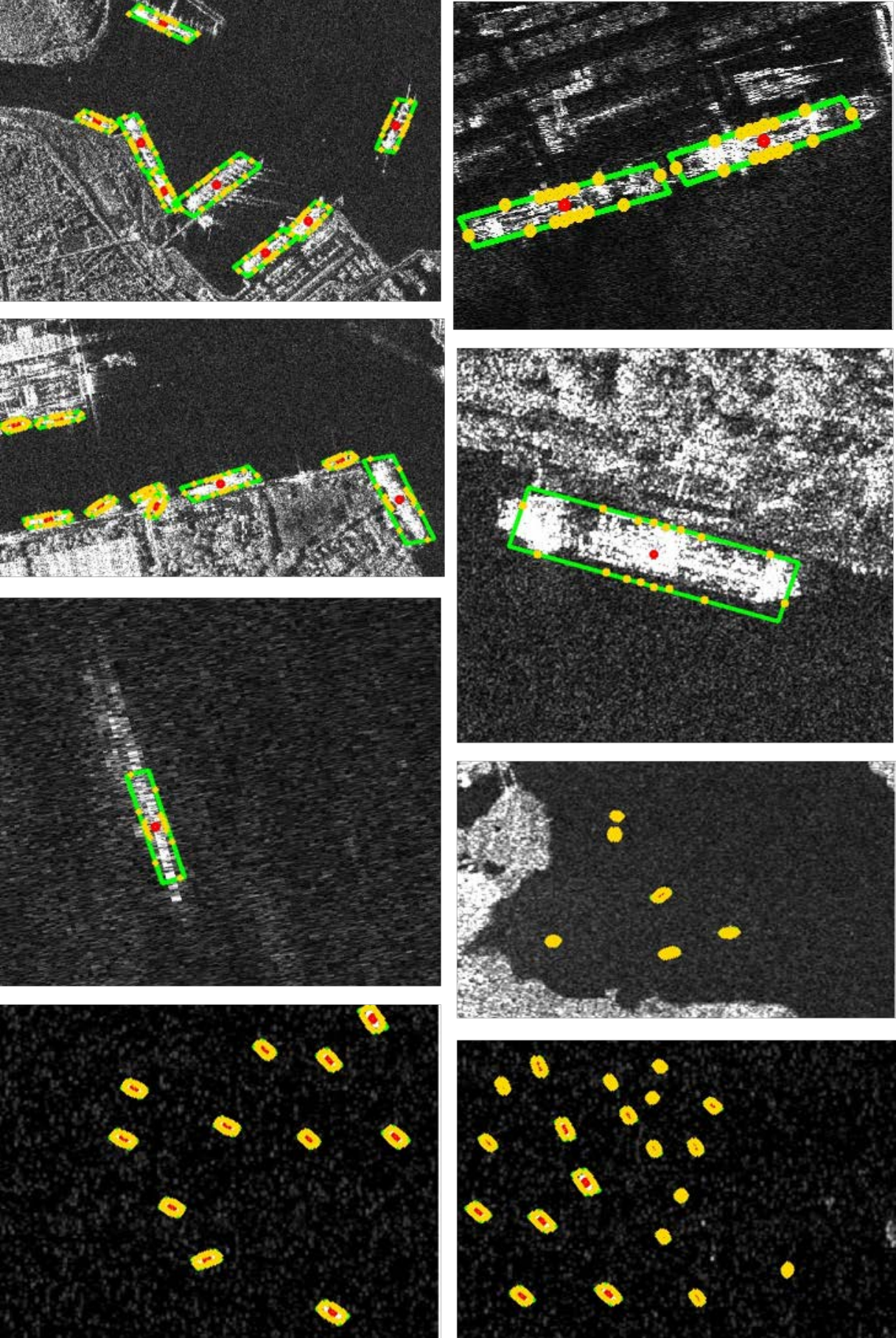}
\caption{The detection results of our method using polar decoding. The red points represent the detected center points of the ship targets; the yellow points denote the boundary points decoded from the polar encodings; and the green rectangles are the minimum bounding boxes of the boundary points, i.e. final detection results.}
\label{fig13}
\end{figure}

In order to visually compare the proposed method with other methods, detection results of different methods from different scenes are given in Fig. \ref{fig12}. Fig. \ref{fig12}(a) shows the ground truth, Fig. \ref{fig12}(b)-(f) give the detection results of our method, BBAVectors, ROITransformer, FRCNN-OBB and RetinaNet-OBB, respectively. From Fig. \ref{fig12}(b), we can see that the proposed method detects the ship most accurately compared with other methods in the inshore scenes, with fewer false alarms and missed detections. To be specific, there is a false alarm in the fourth row and a missed detection in the fifth row. In the last SAR image, there are relatively less false alarms and missed detections, showing more accurate detections. It can be seen from Fig. \ref{fig12}(c) that BBAVectors has more false alarms from the second row to the fourth row than our method. In the fifth image, there exists an inaccurate ship prediction. And in the offshore scene, there are more missed detections. In Fig. \ref{fig12}(d), for ROITransformer, there are false alarms in the first and second images, two missed detections in the fifth image, and some missed detections in the offshore scene. The detection results of FRCNN-OBB in Fig. \ref{fig12}(e) shows a lot of false alarms in all inshore scenes. In the offshore scene, the land clutter is mistakenly detected as the ship target, and some missed detections occur. For the detection results of RetinaNet-OBB in Fig. \ref{fig12}(e), many false alarms and inaccurate detections occur in the inshore scenes. In the offshore scene, there appear many false alarms, showing unsatisfactory detection performance. To summarize, our method achieves more accurate detection results than other methods in both inshore and offshore scenes, which verified the effectiveness of our method.

Fig. \ref{fig13} shows several detection results of our proposed method in the inshore and offshore scenes, where the red points denote the ship centers, the yellow points are decoded boundary points, and the green boxes represent the detection results by finding the minimum bounding boxes of the boundary points. It can be observed from Fig. \ref{fig13} that: (1) the proposed polar encoding method is capable of accurately locating different scales of ship targets in different scenes, indicating the effectiveness of our method; (2) for ship targets with large aspect ratios, it is important to the accurately regress several key points to determine the length of the long side of the OBB. We propose to use the IOU-weighted regression loss to guide the training the polar encodings. In this way, the contribution of these key points on the loss function is increased and the detection accuracy is improved; (3) the number of the boundary points N, which is a hyper parameter, is set to 8 in our experiments. As can be shown, eight boundary points can well represent the OBB of the ship targets in the dataset. But in the future, if our method is applied for detecting targets with larger aspect ratios from other datasets, a larger $N$ is needed for accurate OBB representation.

\section{Discussion}

\subsection{The influence of the hyper parameter N}
\begin{figure}[t]
\centering  
\subfloat[]{  
\includegraphics[scale=0.27]{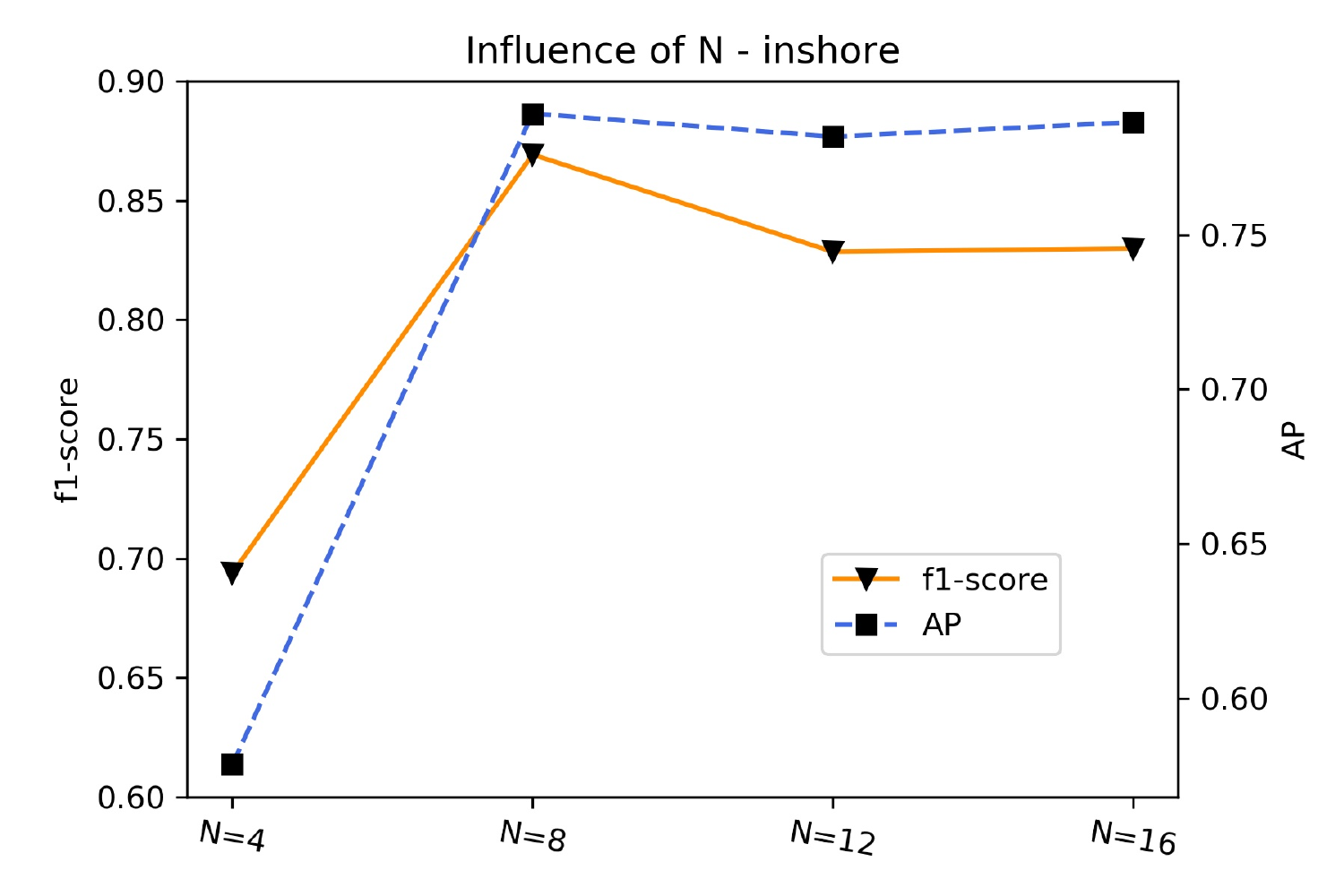}}
\hspace{2pt} 
\subfloat[]{  
\includegraphics[scale=0.27]{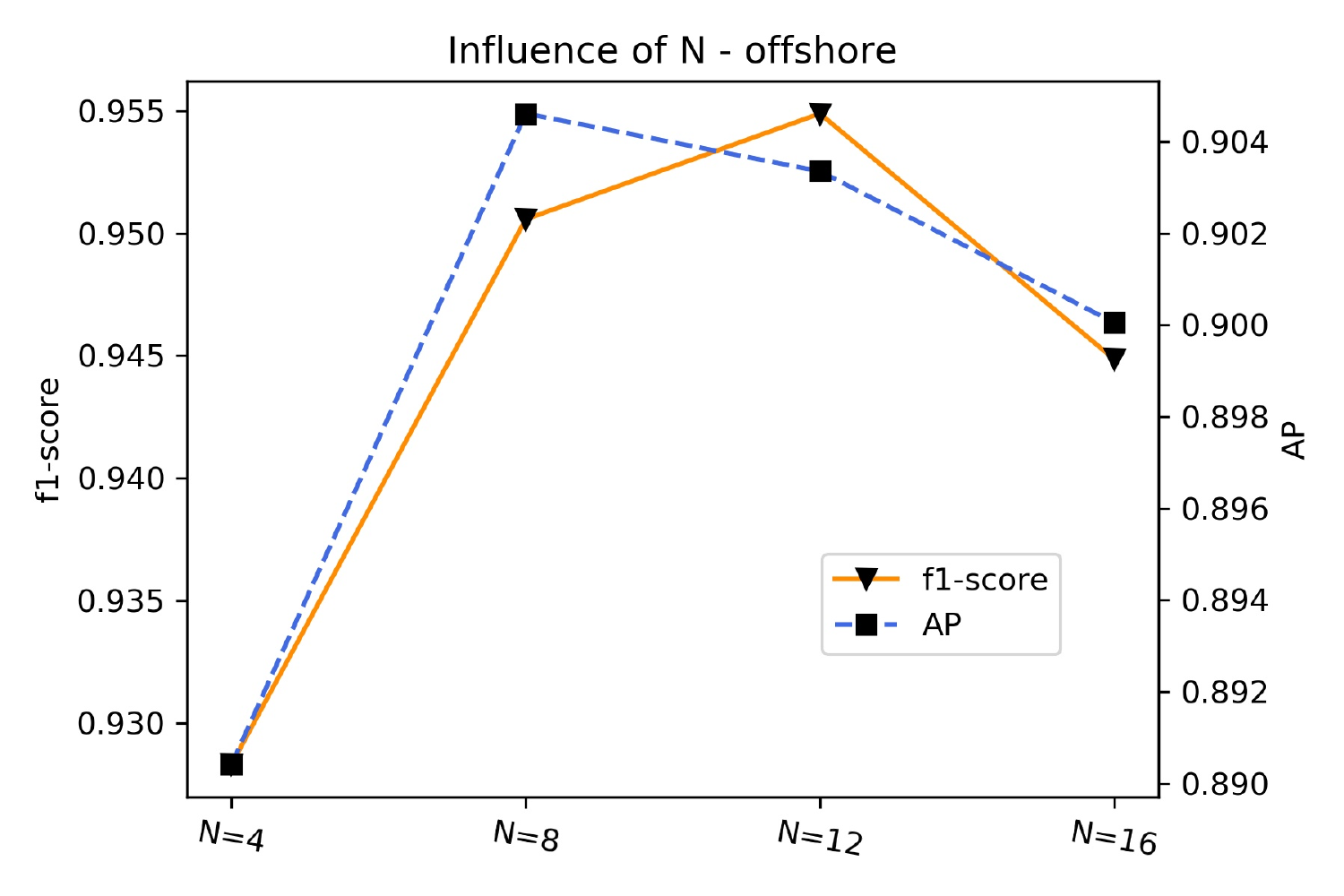}}  
\caption{Detection performance in different scenes under different $N$. (a) AP and F1 in the inshore scenes; (b) AP and F1 in the offshore scenes.}  
\label{fig14}
\end{figure}

The hyper parameter $N$ in the polar encoding process determines the angle sampling rate, which is important for OBB representation. If the sampling rate is too low, the information contained in the sampling points may not adequately represent the shape of the OBB. If the sampling rate is too high, the calculation efficiency will be reduced and the difficulty of training will be increased. Therefore, we test the detection performance of our method under different $N$ values, and the results are shown in Fig. \ref{fig14}. When $N=4$, the detection performance suffers great drop compared to other $N$ values. It is because that the sampling rate is too low to fully represent the shape information of the OBB and guide network training. Among other $N$ values, the detection performance for different scenes is generally robust. For the inshore scenes, the best detection performance is achieved when $N=8$. For the offshore scenes, the AP metric is at the highest when $N=8$ and the F1 achieves the best when $N=12$. According to the above results, $N=8$ is chosen in our experiments for better detection performance and computational efficiency.

\begin{figure}[t]
\centering  
\subfloat[]{  
\includegraphics[scale=0.8]{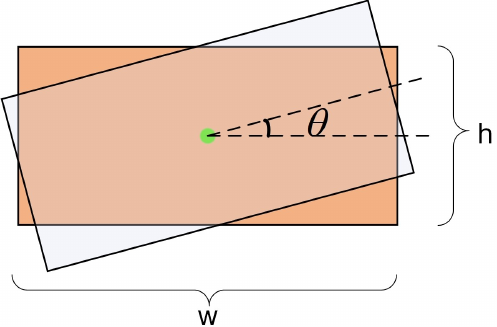}}
\hspace{2pt} 
\subfloat[]{  
\includegraphics[scale=0.6]{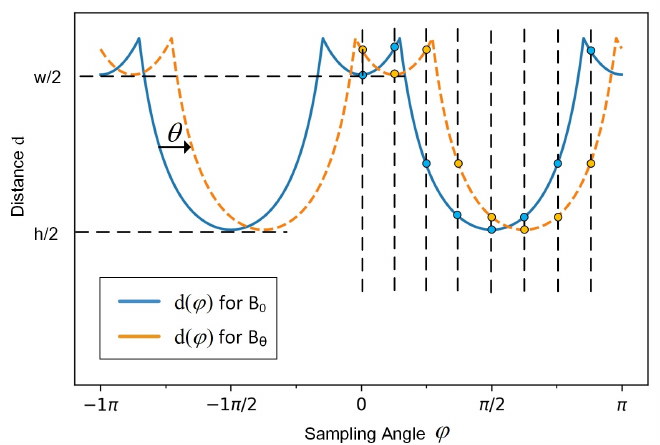}}  
\caption{The diagram of the periodicity of our encoding scheme. (a) Two OBBs with angle error $\theta$; (b) $d(\varphi )$ of the two OBBs.}  
\label{fig15}
\end{figure}

\begin{figure}[t]
\centering  
\subfloat[]{  
\includegraphics[scale=0.3]{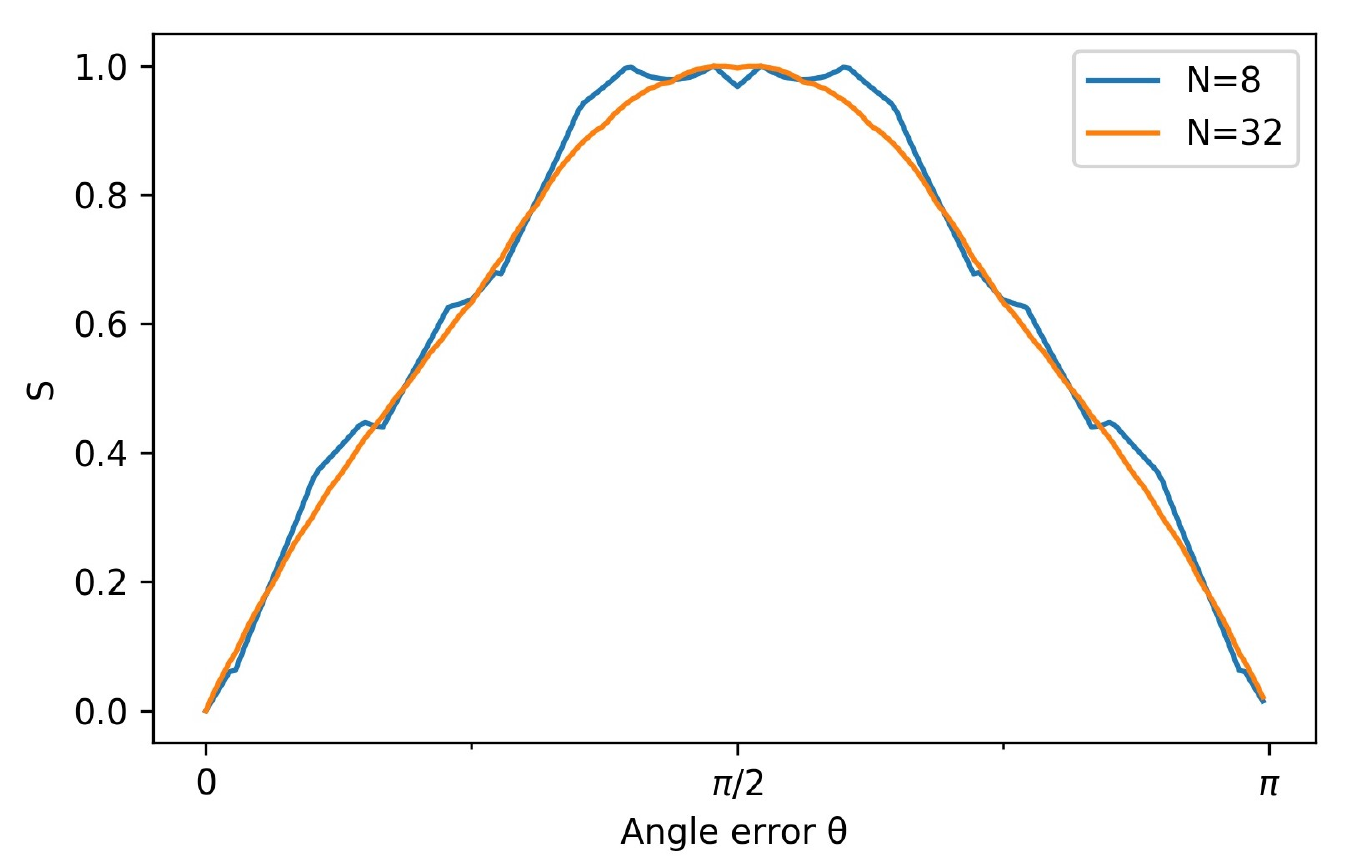}}
\hspace{2pt} 
\subfloat[]{  
\includegraphics[scale=0.3]{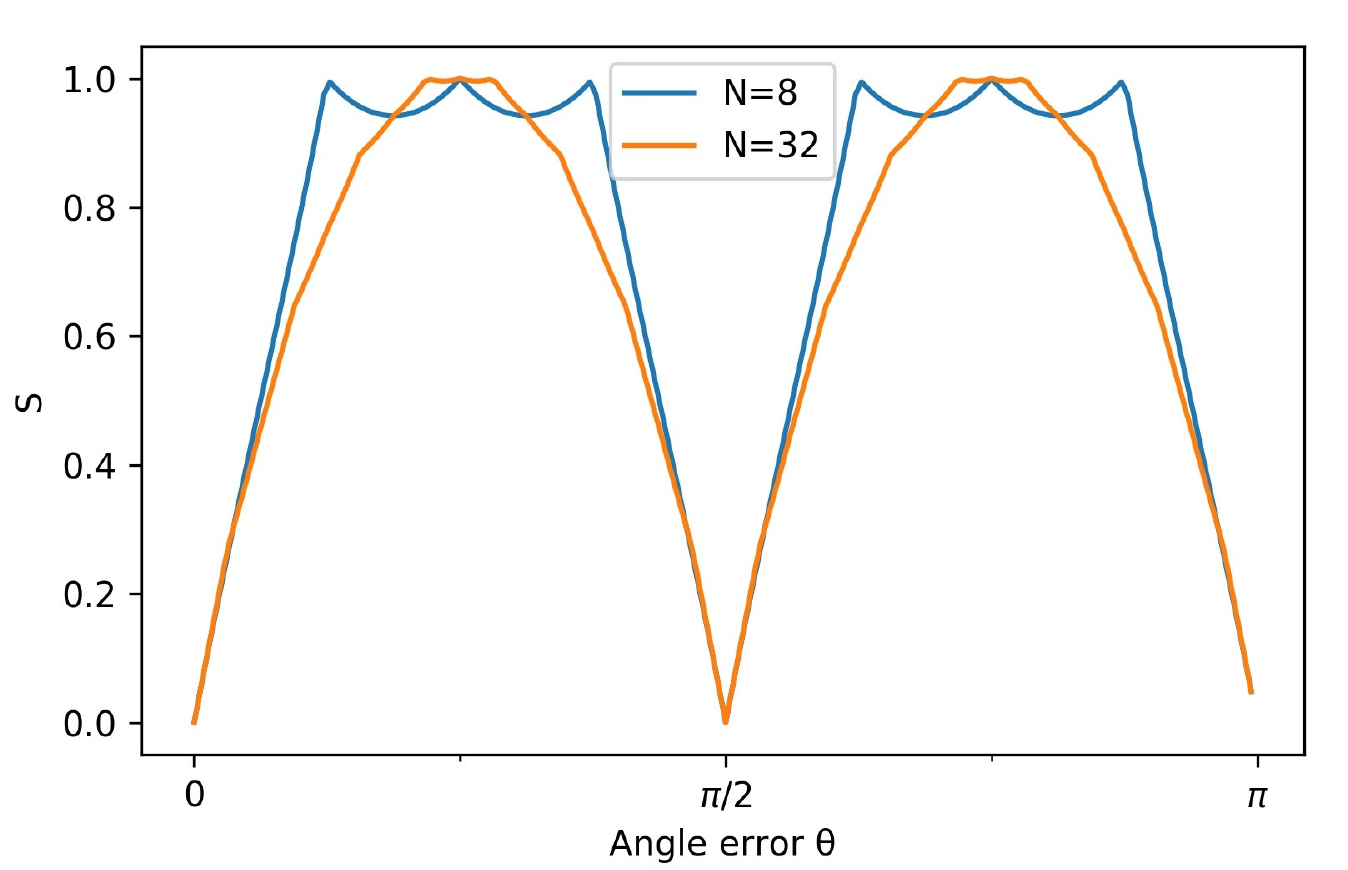}}  
\caption{The curves of $S(\theta )$. (a) $S(\theta )$ for OBB with the aspect ratio of 2; (b) $S(\theta )$ for OBB with the aspect ratio of 1.}  
\label{fig16}
\end{figure}

\subsection{The continuity of our method in the boundary cases}
\label{discussion}
In order to exhibit the continuity of our method in the boundary cases, we can assume the case as shown in Fig. \ref{fig15}(a). Given an OBB $B_{0}$ of the ship target, the height and width of which are $h$ and $w$, respectively. $B_{0}$ rotates clockwise about its center with angle $\theta$, the resulting rotated rectangle is denoted as $B_{\theta}$. In addition, assume that the sampling rate in the polar encoding process is large enough to make the distance $d$ between the center to the boundary of the OBB change approximately continuously with the sampling angle $\varphi $. The function $d(\varphi )$ is plotted in Fig. \ref{fig15}(b), where the blue and yellow curves represent $d(\varphi )$ for $B_{0}$ and $B_{\theta}$, respectively. In fact, the polar encoding process is equivalent to conduct sampling from $d(\varphi )$ in $[0,\pi )$. We can find from Fig. \ref{fig15}(b) that: (1) $d(\varphi )$ is a periodic function, whose period is $\pi$. The two extreme points shown in the figure indicate the height and width information of the OBB; (2) The effect of the rotation angle $\theta$, i.e. the angle error $\theta$, is to make $d(\varphi )$ shift $\theta$; (3) The sum of the absolute differences between the sampling points from the two curves, denoted by $S(\theta )$, indicates the value of the L1-loss. $S(\theta )$ can be calculated as follows:
\begin{IEEEeqnarray}{l}\label{equ19}
    S(\theta )\text{=}\sum\limits_{i=1}^{N}{(\left| {{d}_{0}}({{\varphi }_{i}})-{{d}_{\theta }}({{\varphi }_{i}}) \right|)}
\end{IEEEeqnarray}

Taking the angle error $\theta$ as the independent variable and the sum of the absolute differences $S$ between the sampling points as the dependent variable, the curve of the function $S(\theta )$ is drawn in Fig. \ref{fig16}. Since $S(\theta )$ indicates the value of the regression loss, we can observe how the loss changes with different angle errors. Fig. \ref{fig16}(a) shows the curves of $S(\theta )$ when the aspect ratio of the OBB is 2 and the hyper parameter $N$ is 8, 32. We normalize the values of $S(\theta )$ into $[0, 1]$ for clarity. We can find from Fig. \ref{fig16}(a) that: (1) Within the range of $[0,\pi )$, the loss value first increases from 0 and then decreases to 0. This is ideal because in the actual situations, the overlap between OBBs is the least when the angle error is $\pi/2$, and the overlap between OBBs becomes the largest when the angle error is close to $\pi$. (2) When N is small, the curve is relatively rough. And it becomes smoother when $N$ is in a larger value. But in general, they indicate the same tendency. Fig. \ref{fig16}(b) shows the curves of $S(\theta )$ for OBB with the aspect ratio of 1. It can be seen that the loss value reaches 0 when the angle error is 0, $\pi/2$ and $\pi$, respectively. And the period of $S(\theta )$ is $\pi/2$, which is in accordance with the fact that the square bounding box coincides with itself every $\pi/2$ rotation angle. To sum up, our method can produce corresponding periodic loss functions for different aspect ratios of ship OBBs, so as to overcome the boundary discontinuity problem caused by the periodicity of angle.

\section{Conclusion}
In this paper, we propose an DCNN-based detector using polar encoding and center point detection for arbitrary-oriented ship detection in SAR images. In order to overcome the boundary discontinuity problem caused by the periodicity of angle and the exchangeability of edges, we design the specific encoding and decoding process. In the polar encoding process, the OBB of the ship target is encoded with the help of the polar coordinates of the boundary points. The encoded parameters are trained and regressed end to end. And the polar decoding process is used for restoring the detection results from the encoded parameters. In order to further improve the training of the encoded parameters, we propose an IOU-weighted regression loss, which uses IOU metric to guide network training. Experiments on RSSDD dataset demonstrate that the proposed method can deal with the boundary discontinuity problem better than other encoding schemes, and the IOU-weighted loss can further improve the detection performance. The experimental results also show that our method outperforms other state-of-the-art OBB-based detectors, verifying the effectiveness of our method.


%





\ifCLASSOPTIONcaptionsoff
  \newpage
\fi



\bibliographystyle{IEEEtran}
\bibliography{IEEEabrv,mybibfile}
%



%

\begin{IEEEbiography}[{\includegraphics[width=1in,height=1.25in,clip,keepaspectratio]{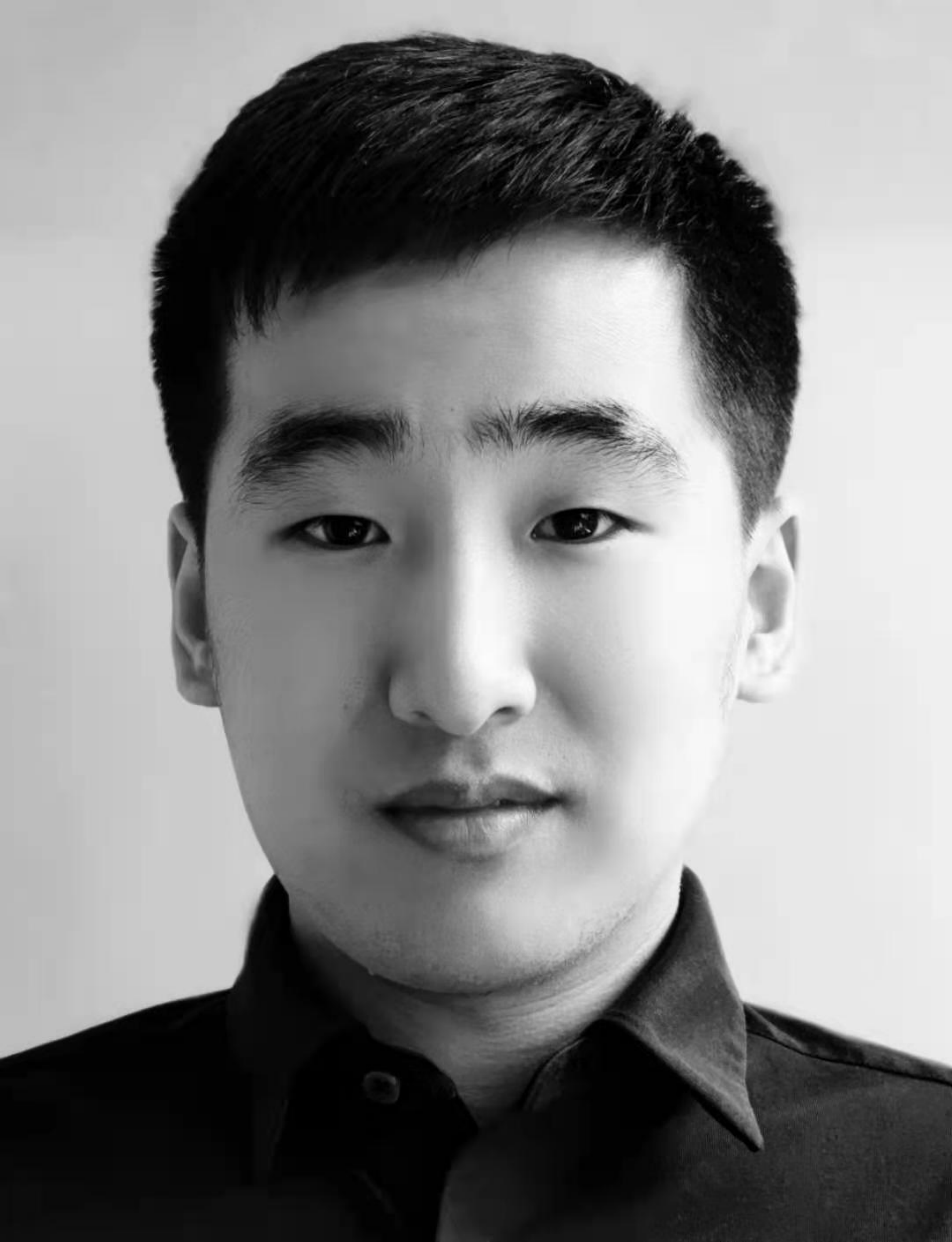}}]{Yishan He}
received the B.S. degree in electronic and information engineering from the Beihang University, Beijing, China, in 2019. He is currently working toward the M.E. degree in information and communication
engineering with Beihang University.

His current research interests include target detection and synthetic aperture radar image processing.
\end{IEEEbiography}

\begin{IEEEbiography}[{\includegraphics[width=1in,height=1.25in,clip,keepaspectratio]{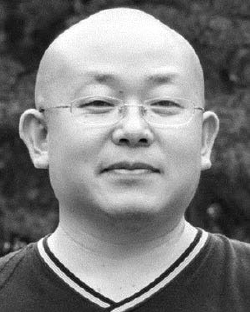}}]{Fei Gao}
received the B.S. degree in industrial electrical automation, the M.S. degree in electromagnetic measurement technology and instrument from the Xi’an Petroleum Institute, Xi’an, China, in 1996 and 1999, respectively, and the Ph.D. degree in signal and information processing from the Beihang University, Beijing, China, in 2005.

He is currently a Professor with the School of Electronics and Information Engineering, Beihang University. His research interests include radar signal processing, moving target detection, and image processing.
\end{IEEEbiography}


\begin{IEEEbiography}[{\includegraphics[width=1in,height=1.25in,clip,keepaspectratio]{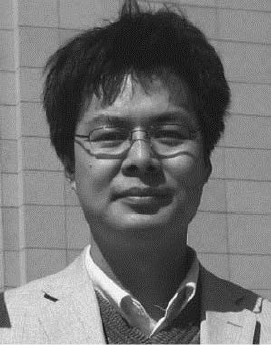}}]{Jun Wang}
received the B.S. degree from the North Western Polytechnical University, Xi'an, China, in 1995, and the M.S. and Ph.D. degrees from the Beijing University of Aeronautics and Astronautics (BUAA), Beijing, China, in 1998 and 2001, respectively.

He is currently a Professor with the School of Electronic and Information Engineering, BUAA. He is interested in signal processing, DSP/FPGA real-time architecture, target recognition, and tracking. His research has resulted in 38 papers in journals, books, and conference proceedings.
\end{IEEEbiography}

\begin{IEEEbiography}[{\includegraphics[width=1in,height=1.25in,clip,keepaspectratio]{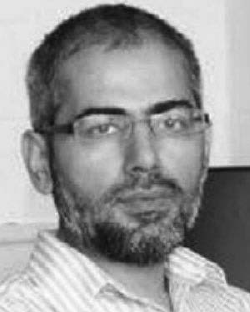}}]{Amir Hussain}
received his B.Eng (highest 1st Class Honours with distinction) and Ph.D degrees, from the University of Strathclyde, Glasgow, U.K., in 1992 and 1997, respectively. 

Following postdoctoral and academic positions at the Universities of West of Scotland (1996-98), Dundee (1998-2000) and Stirling (2000-18) respectively, he joined Edinburgh Napier University in Scotland, UK, in 2018 as founding Director of the Centre of AI and Data Science. Professor Hussain's research interests are cross-disciplinary and industry-led, aimed at developing cognitive data science and AI technologies, to engineer the smart and secure systems of tomorrow.He has (co)authored three international patents and over 400 publications, including 170+ international journal papers, 20 Books/monographs and over 100 Book chapters.
\end{IEEEbiography}

\begin{IEEEbiography}[{\includegraphics[width=1in,height=1.25in,clip,keepaspectratio]{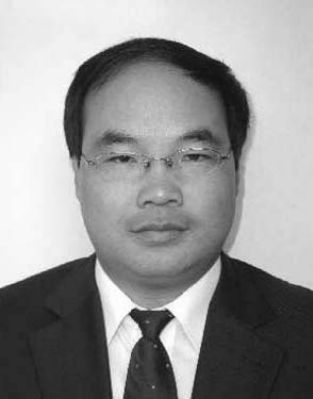}}]{Erfu Yang}
received the Ph.D. degree in robotics in the interdisciplinary area of robotics and autonomous systems from the School of Computer Science and Electronic Engineering, University of Essex, Colchester, U.K. He is currently a Lecturer with the Department of Design, Manufacture and
Engineering Management (DMEM), University of Strathclyde, Glasgow, U.K.

His main research interests include robotics, autonomous systems, mechatronics, manufacturing automation, computer vision, image/signal processing, nonlinear control, process modeling and simulation, condition monitoring, fault diagnosis, multi-objective optimizations, and applications of machine learning and artificial intelligence, including multi-agent reinforcement learning, fuzzy logic, neural networks, bio-inspired algorithms, and cognitive computation. He has over 70 publications in these areas, including over 30 journal papers and ten book chapters.
\end{IEEEbiography}

\begin{IEEEbiography}[{\includegraphics[width=1in,height=1.25in,clip,keepaspectratio]{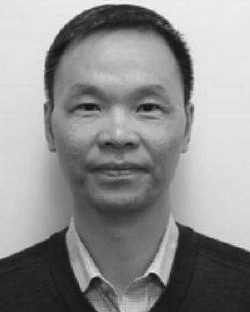}}]{Huiyu Zhou}
received the B.Eng. degree in radio technology from Huazhong University of Science and Technology, Wuhan, China, the M.S. degree in
biomedical engineering from University of Dundee, Dundee, U.K., and the Doctor of Philosophy degree in radio technology, biomedical engineering and computer vision from Heriot-Watt University, Edinburgh, U.K.

He is currently a Professor with the School of Informatics, University of Leicester, Leicester, U.K. He has authored or coauthored widely in medical image processing, computer vision, intelligent systems, and data mining.
\end{IEEEbiography}




\end{document}